\documentclass[journal]{IEEEtran}
\usepackage{graphicx}
\usepackage{float}
\usepackage{subfigure}
\usepackage{amsmath}
\usepackage{amssymb}
\usepackage{booktabs}

\usepackage{threeparttable}
\usepackage{amsfonts}
\usepackage{multirow}

\usepackage{cite}
\usepackage{amsthm,amsmath,amssymb}
\usepackage{mathrsfs}

\usepackage[pagebackref,breaklinks,colorlinks]{hyperref}

\usepackage[capitalize]{cleveref}
\crefname{section}{Sec.}{Secs.}
\Crefname{section}{Section}{Sections}
\Crefname{table}{Table}{Tables}
\crefname{table}{Tab.}{Tabs.}

\ifCLASSINFOpdf
\else
\fi
\begin{document}
%
\title{Mirror Complementary Transformer Network for RGB-thermal Salient Object Detection}
%
%
%

\author{Xiurong Jiang,
        Lin Zhu,
        Yifan Hou,
        Hui Tian,~\IEEEmembership{Senior Member,~IEEE}
\thanks{Manuscript received March 21, 2022. (Corresponding author: Hui Tian.)}
\thanks{X. Jiang, Y. Fan and H. Tian are with the State Key Laboratory of Networking and Switching Technology, Beijing University of Posts and Telecommunications, Beijing, China, 100876 (e-mail: jiangxiurong@bupt.edu.cn; houyifan@bupt.edu.cn; tianhui@bupt.edu.cn).}
\thanks{L. Zhu is with the school of Computer Science, Beijing Institute of Technology, Beijing, China, 100871 (e-mail: linzhu@pku.edu.cn).}}

%
%

\markboth{Journal of \LaTeX\ Class Files,~Vol.~1, No.~1, March~2022}%
{Shell \MakeLowercase{\textit{et al.}}: Mirror Complementary Transformer Network for RGB-thermal Salient Object Detection}
%



\maketitle

\begin{abstract}
RGB-thermal salient object detection (RGB-T SOD) aims to locate the common prominent objects of an aligned visible and thermal infrared image pair and accurately segment all the pixels belonging to those objects. 
It is promising in challenging scenes such as nighttime and complex backgrounds due to the insensitivity to lighting conditions of thermal images. 
Thus, the key problem of RGB-T SOD is to make the features from the two modalities complement and adjust each other flexibly, since it is inevitable that any modalities of RGB-T image pairs failure due to challenging scenes such as extreme light conditions and thermal crossover.
In this paper, we propose a novel mirror complementary Transformer network (MCNet) for RGB-T SOD. 
Specifically, we introduce a Transformer-based feature extraction module to effective extract hierarchical features of RGB and thermal images.
Then, through the attention-based feature interaction and serial multiscale dilated convolution (SDC) based feature fusion modules, the proposed model achieves the complementary interaction of low-level features and the semantic fusion of deep features. 
Finally, based on the mirror complementary structure, the salient regions of the two modalities can be accurately extracted even one modality is invalid. 
To demonstrate the robustness of the proposed model under challenging scenes in real world, we build a novel RGB-T SOD dataset VT723 based on a large public semantic segmentation RGB-T dataset used in the autonomous driving domain. 
Expensive experiments on benchmark and VT723 datasets show that the proposed method outperforms state-of-the-art approaches, including CNN-based and Transformer-based methods. 
The code and dataset will be released later at https://github.com/jxr326/SwinMCNet. 
\end{abstract}


%
\IEEEpeerreviewmaketitle

\section{Introduction}
\label{sec:motivation}

\begin{figure}
\centerline{\includegraphics[width=\columnwidth]{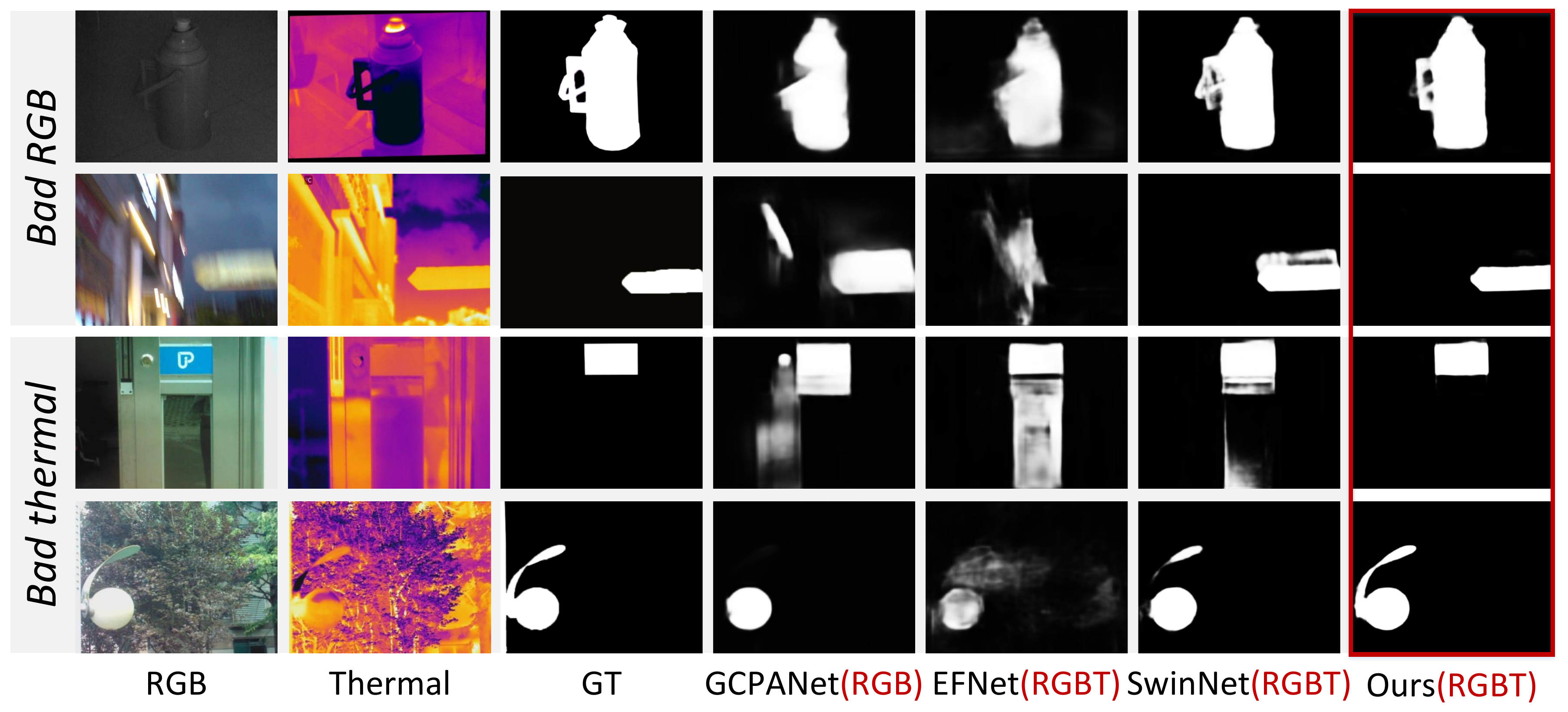}}
\vspace*{-3mm}
\caption{\textbf{Visual illustration of complementary benefits of RGB and thermal modalities.} The above two rows show challenging scenes of low illumination and overexposure, while the bottom two rows shows scenes of complex thermal environments such as thermal crossover. 
The results show that the proposed RGBT SOD model performs better than other representative single-modality SOD method~\cite{chen2020global} and multi-modality SOD methods~\cite{chen2021ef,liu2022swinnet}.}
\vspace*{-4mm}
\label{fig1}
\end{figure}

\IEEEPARstart{S}{alient} object detection focuses on mimicking the attention mechanism in the human visual system (HVS) to locate the most salient objects in a scene~\cite{borji2019salient}, which is applied in various fields such as computer vision, computer graphics and robotics. 
Representative applications include image understanding, semantic segmentation, non-photo-realist rendering, automatic image cropping and human-robot interaction, etc. 
Although significant progress has been made in the SOD field over the past several years, there is still room for improvement in the single modality SOD (i.e., detection on a single RGB input image) when faced with challenging factors, such as low illumination conditions or the background is cluttered in the scenes~\cite{zhou2021rgb,fu2020light}. To break through the performance bottleneck, researchers try to introduce additional supplementary knowledge to cope with challenging scenes of SOD, such as depth maps~\cite{zhou2021rgb}, thermal maps~\cite{tu2020rgbt} or light field data~\cite{fu2020light}.

In the field of multi-modal SOD, the depth map are introduced earlier by researchers as a complementary information to RGB image, namely RGB-D SOD. The value of each pixel in the depth map represents the distance of that point from the camera in the scene. Thus the depth map usually contains rich spatial position and structural information about the salient objects. It is often used as an auxiliary modality to handle challenging scenes such as low contrast and complex backgrounds~\cite{zhao2021rgb,fan2020bbs,chen2021ef}. However, since depth map is easily disturbed by low illumination and occlusion, which limits the practical applications~\cite{zhao2021rgb}.

Thermal infrared cameras convert temperature differences into images by capturing infrared radiation from all objects in nature with temperatures above absolute zero. Its imaging mechanism does not depend on lighting conditions, thus the combination of RGB and thermal modalities has natural advantages in challenging scenes such as bad weather, night and so on. RGB-T provides new ideas for many challenging computer vision tasks~\cite{huo2021efficient}, for example, RGB-T tracking, RGB-T crowd counting, and RGB-T person re-identification. 
In this paper, we focus on exploring the salient objection detection task in an aligned visible and thermal infrared image pair, which is named RGB-T SOD. 

In fact, RGB images contain rich color and texture features, in most cases the objects are more salient relative to the background. For thermal images, the infrared radiation from the same object tends to be uniform or varies very subtle, while the temperature difference between different objects is relatively large (except for thermal crossover occurs). It can be said that the same object is always homogeneous and the edge of objects is usually clear in the thermal imaging of a scene by an infrared camera. 

As shown in Fig.~\ref{fig1}, in the scenes under low illumination or image blur, traditional cameras cannot capture enough details of the objects (see the top two rows). It is difficult for RGB-based SOD methods (e.g., GCPANet~\cite{chen2020global}) to accurately segment salient objects in this case. The RGB-D SOD methods usually regard the depth maps as auxiliary information of RGB images, and use a lightweight network to learn the features of objects from depth maps. 
Due to different data characteristics, it could not obtain satisfactory results by simply applying the RGB-D SOD model to the RGB-T SOD task (e.g., EFNet~\cite{chen2021ef}). As a recent RGB-T SOD model, SwinNet~\cite{liu2022swinnet} performs well in most scenes, however, it is still not accurate enough in some challenging scenes (e.g., bad thermal scenes).
The last column in Fig.~\ref{fig1} illustrates that our model can make full use of the complementary benefits of RGB and thermal modalities, thus obtain satisfactory results.

To make the features from the two modalities complement and adjust each other flexibly, we design a mirror complementary network (MCNet) based on the complementary advantages of RGB and thermal images (see Fig.~\ref{fig2}). 
Inspired by the successful use of Transformer networks in the field of computer vision~\cite{liu2021swin,liu2022swinnet}, we study the RGB-T SOD method based on Transformer and CNN hybrid structure. 
The hybrid structure (e.g., Swin Transformer~\cite{liu2021swin}) combines the long-range dependency merit of Transformer and the locality and hierarchy advantages of CNN, and its computational complexity is limited to a linear function of image size through shifted window operation, so it is very suitable for pixel-level prediction tasks with high image resolution. 
In this paper, we extract the hierarchical features of RGB and thermal modalities respectively based on a Transformer feature extractor, and model the complementary relationship of saliency information implied in cross modalities apply the locality merit of CNN.

The idea of mirror complementary is similar to full-duplex or bi-directional strategies utilized in video or RGB-D SOD~\cite{ji2021full,zhang2021bts}, we design the overall structure of MCNet according to the characteristic of RGB-T data.
The two streams of MCNet have fully equivalent encoding and decoding structures, and use a set of complementary grayscale labels to supervise the salient features of the two modalities respectively. 
Specifically, the complementary grayscale labels control the RGB stream to focuses on the skeleton region of the salient objects and the thermal stream to focuses on the contour of the salient objects. 
On the one hand, different color blocks of the same object can easily be misjudged as different objects in the RGB image, while the thermal image is likely to be consistent. On the other hand, thermal image has single color and low contrast between objects and background. 
Therefore, as shown in Fig.~\ref{fig2}, we embed an attention-based feature interaction module between the low-level features of the two modalities, which introduces thermal information into the skeleton branch dominated by RGB to avoid the separation of the same object and maintain the integrity of the salient region.
Also, RGB information is introduced into the contour branch dominated by thermal to enable the network to capture the saliency of the objects and filter out irrelevant background clutters. 
Furthermore, in order to fuse the complementary semantic features of the two modalities and suppress the saliency bias, we introduce a serial multiscale dilated convolution (SDC) based feature fusion module to make the model focus on the common salient regions and obtain more accurate segmentation.

In summary, this paper makes three major contributions:

\begin{itemize}
\item We design a mirror complementary RGB-T SOD network (MCNet) with Transformer and CNN hybrid architecture. By combining the Transformer-based feature extraction module and CNN-based feature interaction and fusion modules, the proposed model can effective utilize the complementary information of the two modalities, thus outperforming the state-of-the-art RGB-T SOD models.

\item An attention-based feature interaction module and a SDC-based feature fusion module are proposed to achieve the complementary fusion of low-level features and semantic features between RGB and thermal modalities. The two modalities can adjust each other and complement flexibly, which significantly improves the predicted saliency map.

\item We build a more challenging RGB-T SOD dataset VT723 and make it available to the research community. Most of the scenes of this dataset contain challenges common in the real world such as low illumination, complex background and so on. 
Extensive experiments on three benchmark RGB-T SOD datasets and VT723 show that the proposed model performs well in different challenging scenes.

\end{itemize}

The rest of this paper is organized as follows: the related works are introduced in Sec. II, and Sec. III illustrates the proposed MCNet. Expensive experiments on benchmark datasets and the proposed V723 dataset are conducted in Sec. IV. Finally, we conclude our work in Sec. V.

\section{Related Work}

\subsection{RGB salient object detection}

The early non-deep SOD models rely on low-level features and certain heuristics~\cite{borji2019salient} (e.g., color contrast, texture, structure and background prior). With the successful application of deep learning technology in CV field, various deep learning SOD models have been proposed. 
The deep model based on multi-layer perceptron (MLP)~\cite{zhao2015saliency,wang2015deep} first showed better performance than traditional scheme. 
Later, the Fully Convolutional Networks (FCN) with end-to-end pixel-level operation was adopted to further improve SOD efficiency and performance, and became the mainstream of SOD architecture~\cite{chen2020global,wei2020label}. According to the basic idea of feature representation, these methods are summarized into three representative categories: edge-focus or boundary-focus~\cite{zhao2019egnet}, multiscale learning~\cite{pang2020multi} and integrity learning~\cite{zhuge2021salient}.

Although deep learning based algorithms have greatly promoted the development of SOD, the existing methods based on single image source RGB still face some challenges to extremely complex scenes. This motivates researchers introduce additional information for SOD task.

\subsection{RGB-D salient object detection}

Existing CNN-based RGB-D SOD methods~\cite{zhao2020single,fu2020jl,chen2021ef,zhang2021uncertainty} can be divided into three categories according to the mode of feature extraction and cross-modal fusion. 
The first category uses input fusion to directly combine the original data of the two modalities. Some researches ~\cite{zhao2020single,fan2020rethinking} concatenated RGB and depth across channel dimension as the raw input of the encoder, and Chen \emph{et al.}~\cite{chen2021rgb} introduces 3D CNN to extract saliency features of the high-dimensional fusion modality. 
The second category~\cite{fu2020jl,fu2021siamese} adopt a Siamese network with shared weights for the RGB/depth stream as an encoder and conduct independent feature extraction of the two modalities. 
The third category extract features from RGB and depth maps independently by using a two-stream network, according to the fusion stage, it can be further divided into result fusion~\cite{wang2019adaptive,chen2020rgbd} and feature fusion~\cite{chen2021ef,li2021hierarchical,fan2020bbs,zhou2021irfr,zhou2021ccafnet,zhang2021rgb}. 
The result fusion generate the final saliency map by fusing the respective predictions of the two modalities. 
Feature fusion usually design complex fusion module based on the progressive fusion mechanism of encoder and decoder,  
and this kind of model gradually becomes the mainstream structure of RGB-D SOD~\cite{chen2021ef,li2021hierarchical}.

\begin{figure*}
\centerline{\includegraphics[width=2\columnwidth]{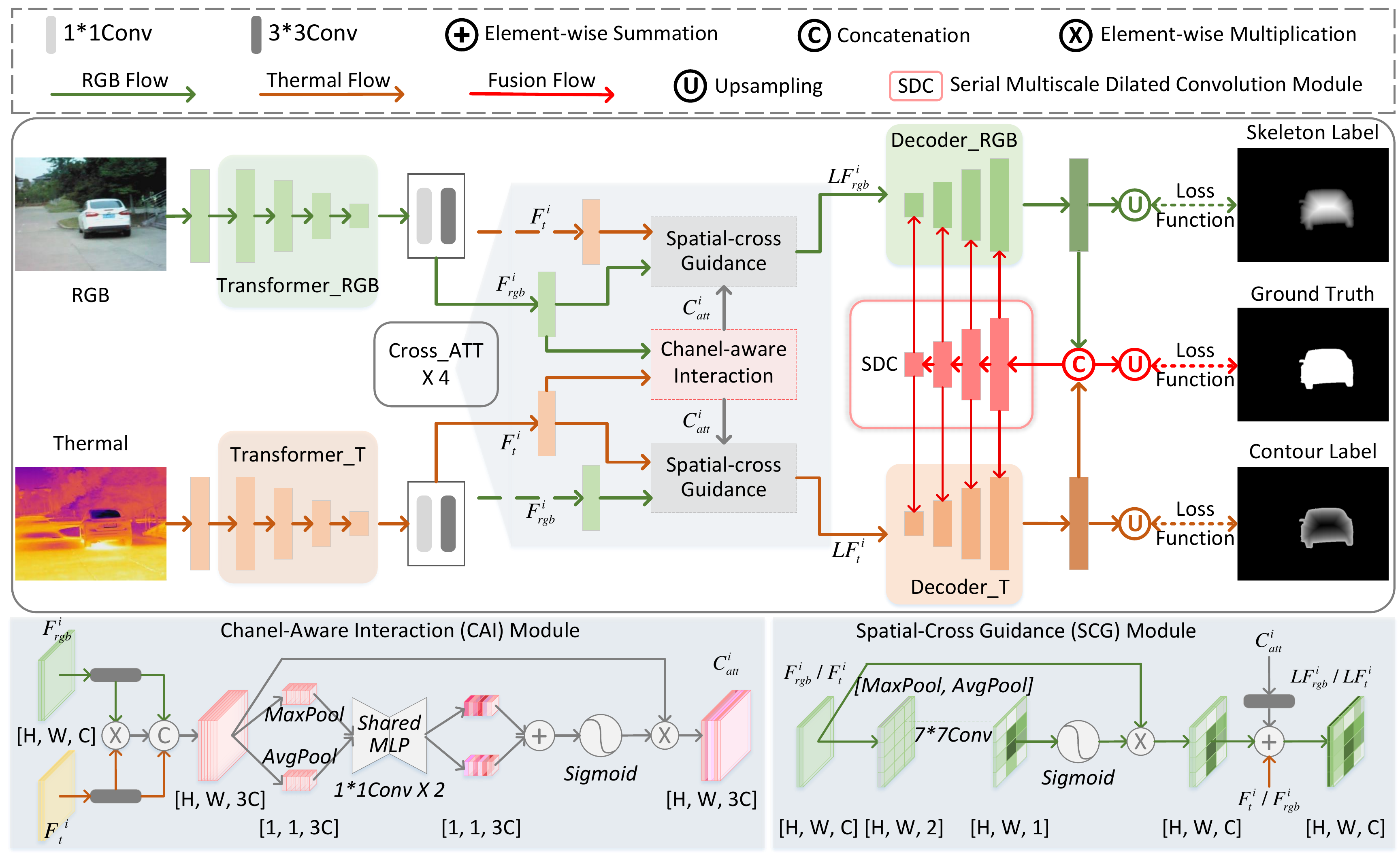}}
\vspace*{-2mm}
\caption{\textbf{The pipeline of mirror complementary network (MCNet)}. MCNet uses a Transformer-based two-stream structure to extract hierarchical features of RGB and thermal modalities, and adopts a complementary set of labels to supervise them respectively. Specifically, the encoders are based on Transformer backbone, followed by four-layer CNN-based decoders.
Through the attention-based feature interaction module ($Cross\_{ATT}$) and SDC-based feature fusion module (SDC), the proposed model achieves the complementary fusion of low-level features and semantic features between RGB and thermal modalities.}
\vspace*{-4mm}
\label{fig2}
\end{figure*}

\subsection{RGB-T salient object detection}
\label{sec:RGB-T SOD}

Early RGB-T SOD methods mostly use graph-based techniques and design bottom-up or top-down models to learn cross-modal feature representations. 
Wang \emph{et al.}~\cite{wang2018rgb} create the first benchmark RGB-T SOD dataset called VT821 and propose a graph-based multi-task manifold ranking algorithm to fuse RGB and thermal data. 
Tu \emph{et al.} adopted multi-mode multi-scale manifold ranking~\cite{tu2019m3s} and cooperative graph learning algorithm~\cite{tu2019rgb} to achieve cross-modal SOD, and they also built a more challenging dataset VT1000 in~\cite{tu2019rgb}. 

In later works, Tu \emph{et al.}~\cite{tu2020rgbt} contributed a large dataset VT5000, which provides benchmark training data for deep learning-based RGB-T SOD. With this dataset, they proposed a baseline deep learning method based on two-stream structure. 
Similar to~\cite{tu2020rgbt}, later CNN-based methods~\cite{huo2021efficient,zhou2021ecffnet,tu2021multi,wang2021cgfnet} mostly adopted feature fusion in two-stream structure to achieve cross-modal fusion for accurate RGB-T SOD. 
Zhou \emph{et al.}~\cite{zhou2021ecffnet} used a bilateral inversion fusion module to bilaterally fuse the foreground and background information. 
To prevent the information of the two modalities from over-influencing each other, Tu \emph{et al.}~\cite{tu2021multi} only performed fusion in the decoding stage. 
Wang \emph{et al.}~\cite{wang2021cgfnet} designed step by step fusion of cross-modal from the perspective of mutual guidance of two modalities. 
The key of RGB-T SOD is to exploit the correlations between RGB and thermal images and fuse these two modalities effectively to accurately segment the common salient regions.

\subsection{Vision Transformer for SOD}

Inspired by the breakthroughs from Transformer~\cite{vaswani2017attention} networks in Natural Language Processing (NLP) domain, researchers applied it to computer vision tasks and achieved remarkable results~\cite{han2020survey,khan2021transformers}. 
Dosovitskiy \emph{et al.}~\cite{dosovitskiy2020image} proposed ViT model based on transformer for the first time in large-scale supervised image classification tasks. 
Wang \emph{et al.}~\cite{wang2021pyramid} designed a progressive shrinking pyramid Transformer named PVT, and Liu \emph{et al.} ~\cite{liu2021swin} proposed Swin Transformer with sliding window operation and hierarchical design. 
As backbone of dense prediction, these two structures can combine numerous detection methods and have good performance in various image/video tasks. 

Subsequently, researchers apply Transformer to SOD tasks to further improve the detection performance. 
Ren \emph{et al.}~\cite{ren2021unifying} and Zhu \emph{et al.}~\cite{zhudftr} applied the pure Transformer-based encoder on single modality SOD, while some researches~\cite{liu2022sdetr,qiu2021boosting} adopted PVT~\cite{wang2021pyramid} and CNN hybrid structure for feature extraction and saliency map prediction. 
Some researches~\cite{liu2021visual,liu2021tritransnet} modeled the long-range characteristics of RGB and depth respectively based on pure Transformer architecture. And others~\cite{pang2021transcmd,wang2021mtfnet,fang2022grouptransnet} combined CNN and Transformer into hybrid architecture, then used the similarity-based attention mechanism of Transformer to achieve accurate RGB-D SOD. 
Liu \emph{et al.}~\cite{liu2022swinnet} uses Swin Transformer~\cite{liu2021swin} encoder to extract features of RGB and thermal/depth images, and then adopt edge guidance to achieve cross-modal fusion for SOD. 

As discussed in previous section, the characteristic of RGB-T data is different from that of RGB-D data. Therefore, it is necessary to design an effective SOD network based on RGB-T data. 
In this work, we propose a Transformer-CNN hybrid network to effective utilize the complementary information of thermal and RGB modalities, which performs well in a variety of challenging scenes.

\section{Methodology}

\subsection{Motivation}
Generally speaking, RGB modality contains rich color and texture information, so the objects are more significant; while the thermal modality contains clear and continuous edge information.
However, the RGB or thermal data in RGB-T image pairs is not always helpful for SOD due to the influence of complex imaging conditions.
For example, the noisy and low contrast RGB modality caused by low light conditions (e.g., the first row in Fig.~\ref{fig1}) and the failure of thermal modality caused by thermal crossover (e.g., the fourth row in Fig.~\ref{fig1}). 
The imaging mechanism of thermal modality determines that RGB-T data is significantly different from RGB-D data~\cite{zhao2021rgb}, since the former allows bad RGB imaging conditions (RGB may not contain much information) but the latter is usually based on high quality RGB image~\cite{zhou2021rgb,fan2020bbs,chen2021ef}. 
Therefore, the key of the RGB-T SOD task is how to extract features conducive to salient objects representation based on the potential cues of the two modalities, and make them mutually adjustable and flexible complementary. 
To reflect challenging scenes such as extreme illumination and complex backgrounds that are common in the real world, we build an RGB-T SOD dataset containing 723 pairs of RGB-thermal images based on autonomous driving scenes (see Sec.~\ref{sec:datasets}).

\subsection{Overview of the proposed network}

We model the common saliency of RGB and thermal modalities by a symmetric two-stream network supervised by a pair of complementary labels. The RGB images emphasize the overall saliency of objects with its rich color and texture information, while the thermal images can help to obtain better spatial consistency and contour localization. Specifically, the hierarchical features of two modalities are first constructed based on Transformer-based feature extraction module, then an attention-based feature interaction module are designed to perform channel-aware interaction and spatial-cross interaction on the feature maps. Finally, a serial multiscale dilated convolution module is proposed to generate the refined fusion result.
Fig.~\ref{fig2} shows the overview of the proposed model, the details are described in the following sections.

\subsection{Transformer-based feature extraction module}

In this section, we design a Transformer-based feature extraction module.
The hierarchical features of two modalities are constructed based on two independent Swin Transformer~\cite{liu2021swin} backbone, then the features of last four layers are used for further feature interaction and fusion.
By utilizing self-attention based on non-overlapping local windows and cross-window mechanisms, the locality and hierarchy of CNN and the long-range dependencies of Transformer are introduced simultaneously.

\begin{figure}
\centerline{\includegraphics[width=\columnwidth]{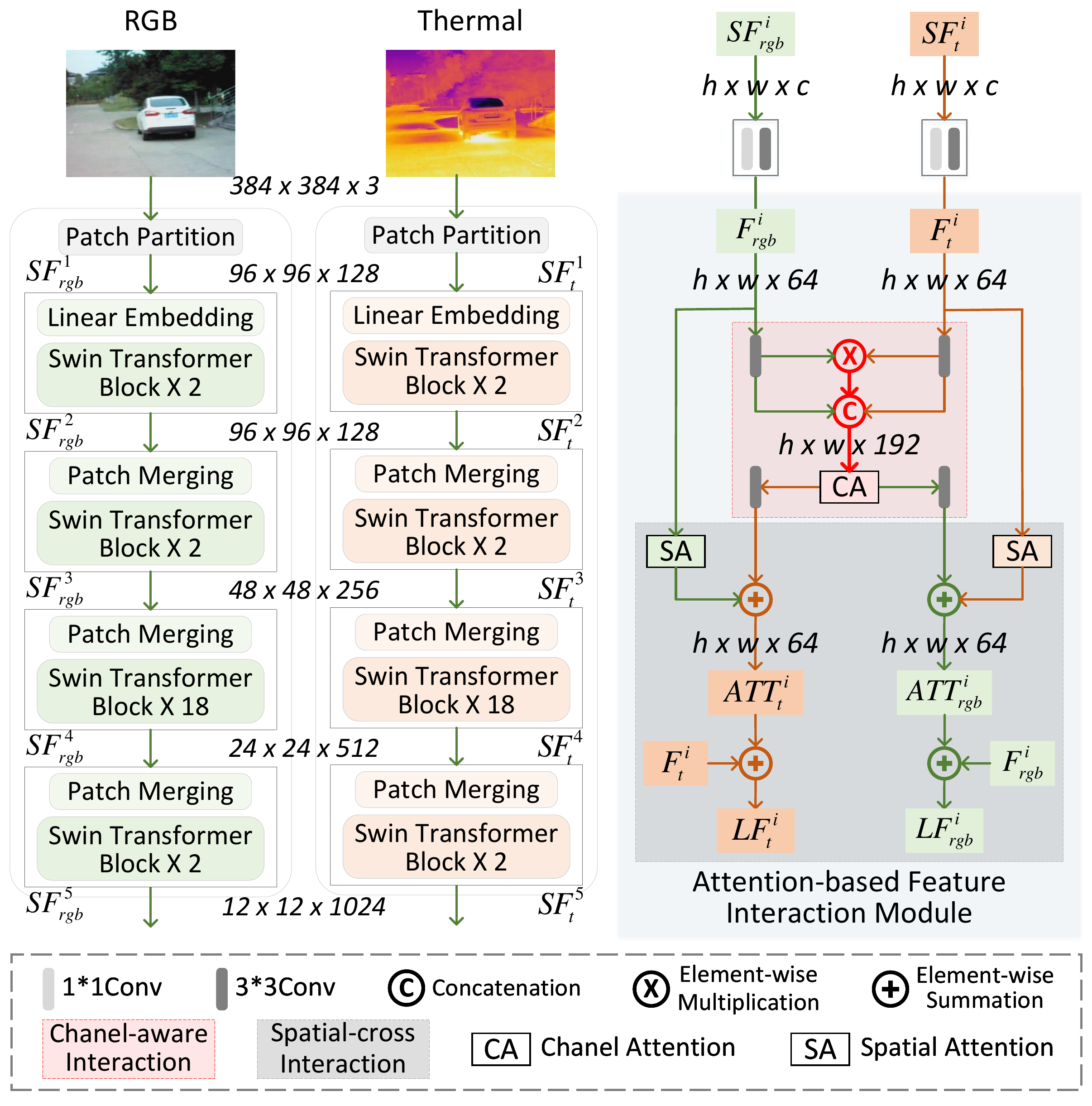}}
\vspace*{-1mm}
\caption{\textbf{The proposed attention-based feature interaction module.} We use Swin-B as the backbone encoder to extract the multi-scale features of the two modalities as $SF^i_{rgb}$ and $SF^i_{t}$, and then two groups of features of the same size are fed into the attention-based interaction module to obtain the low-level interaction features of two modalities as $LF^i_{rgb}$ and $LF^i_{t}$. Each layer of the attention-based interaction module includes a channel-aware interaction module and a spatial-cross interaction module.}
\vspace*{-3mm}
\label{fig3}
\end{figure}

Considering the trade-off between model performance and computational efficiency, as shown in Fig.~\ref{fig3}, we use the pre-trained Swin-B as the backbone encoder, which accept the input size of 384*384. Firstly, the input is segmented into a series of non-overlapping patches. Then, the model adopts a hierarchical design, consisting of 4 stages in total. The channels of the input feature map of each stage are doubled and the resolution is halved, so as to expand the receptive field layer by layer. At this point, we can obtain two groups of feature maps from the two modalities containing five different resolutions, which are respectively denoted as: $SF_{rgb} = \{SF^i_{rgb}|i = 1,2,3,4,5\}$ and $SF_t = \{SF^i_{t}|i = 1,2,3,4,5\}$. 
Compared to high-level features, low-level features contribute less to the performance of deep polymerization methods~\cite{wu20193907}. On the other hand, the larger spatial resolution of low-level features means more computation consumption. 
Therefore, in order to make full use of the global long-range dependency features modeled by Transformer and avoid excessive computing costs, we retain the features of the last four layers (that is, the outputs from the four stages of Swin-B) and discard the lowest features $SF^1_{rgb}$ and $SF^1_{t}$. 

\begin{figure}
\centerline{\includegraphics[width=\columnwidth]{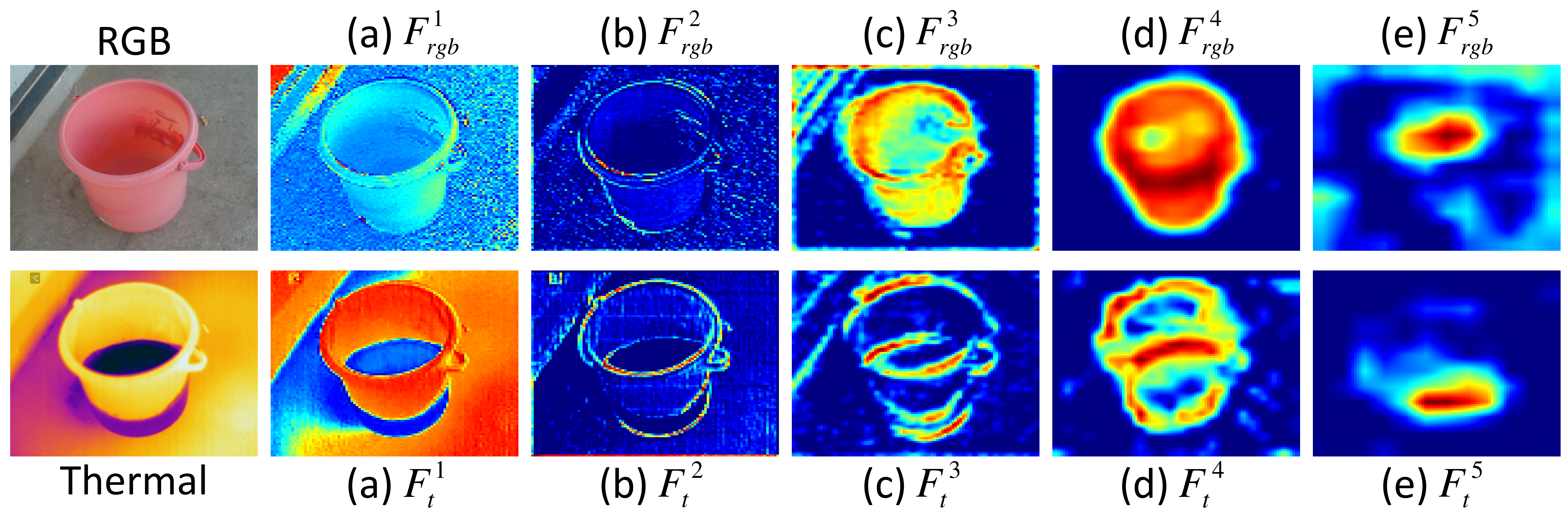}}
\vspace*{-2mm}
\caption{\textbf{The original image pair and five-level feature maps from Swin-B.} From (a) to (e) are the feature maps from different levels of the encoder, respectively. The feature maps in (a) are similar to the origin inputs with background noise, while the deep features such as egde and body details are mainly extracted in (b) - (e).}
\label{fig4}
\end{figure}

In Fig.~\ref{fig4}, we show an example of hierarchical features outputted from Swin-B. As can be seen, the lowest features (a) is similar to the original input because it has not yet passed through Transformer blocks. On the one hand, it contains more background clutter; On the other hand, the edge information of the object is reserved in the following two layers of features (b) and (c) (especially (b)). 
The ablation experiments in Sec.~\ref{sec:ablation} about the hierarchical features of Swin-B demonstrate that the last four levels contain more valid information that can represent salient objects.

\subsection{Attention-based feature interaction module}
As analyzed above, the low-level features from Transformer backbone capture continuous boundary and global details but also with noise, while the high-level features capture semantic feature of salient objects but fail to extract boundaries. 
In order to distill out effectively squeezed cues from two modalities and adjust them each other, we propose a novel attention-based feature interaction module followed the feature extraction module.

As show in Fig.~\ref{fig3}, the hierarchical features $SF_{rgb}$ and $SF_t$ outputted by Swin Transformer blocks are uniformly squeezed into 64 channels through two concatenated convolution operations with kernel size 3$\times$3 and 1$\times$1. We denote these two groups of features as: $F_{rgb} = \{F^i_{rgb}|i = 2,3,4,5\}$ and $F_t = \{F^i_{t}|i = 2,3,4,5\}$. Then the proposed attention-based feature interaction module performs channel-aware interaction and spatial-cross interaction on each set of feature maps with the same resolution. 
We get two sets of attention interaction maps by: 
\begin{equation}
\label{eq1}
Att^i_{rgb} = S_{att}(F^i_{t})+Conv^{3\times3}(C_{att}(F^i_{fuse})),
\vspace*{-2mm}
\end{equation}

\begin{equation}
\label{eq2}
Att^i_{t} = S_{att}(F^i_{rgb})+Conv^{3\times3}(C_{att}(F^i_{fuse})),
\end{equation}
where $C_{att}(\cdot)$ and $S_{att}(\cdot)$ denote the channel attention and spatial attention~\cite{woo2018cbam}, respectively. 
$F^i_{fuse}$ is the shared channel-aware feature of the two modalities, and the detailed structure is shown at the bottom of Fig.~\ref{fig2}.

\begin{figure}
\centerline{\includegraphics[width=\columnwidth]{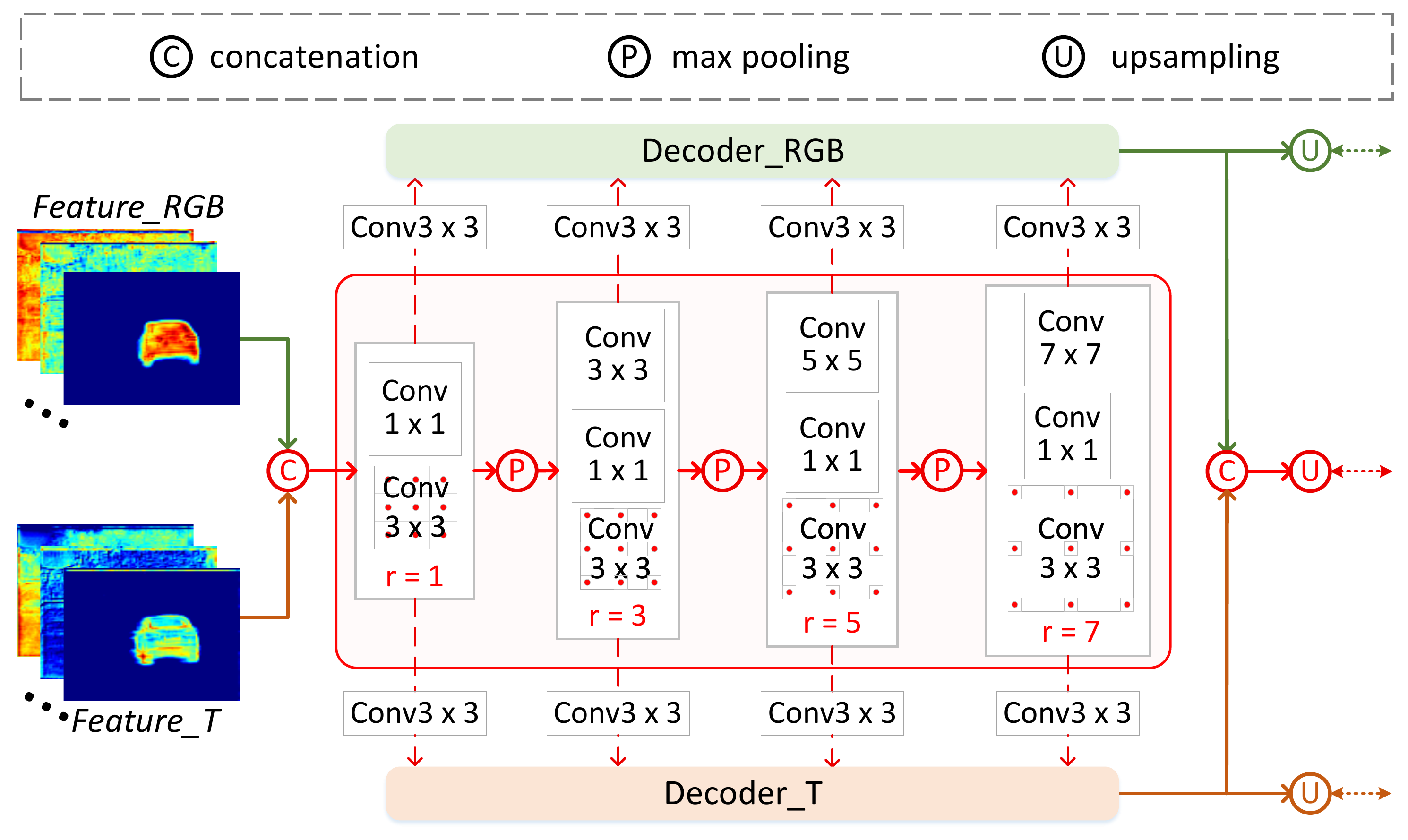}}
\vspace*{-2mm}
\caption{\textbf{The serial multiscale dilated convolution (SDC) module.} SDC module takes the concatenated features of two modalities as inputs, through the cascade of dilated convolution blocks with different expansion rates to expand the receptive field. The low-level interaction features from the two modalities are further integrated and fused.}
\vspace*{-4mm}
\label{fig5}
\end{figure}

\textbf{Channel-Aware Interaction (CAI) module.}
The CAI module is designed to select the saliency consistency features of channel correlation between two modalities ($C_{att}^i$ as shown at the bottom of Fig.~\ref{fig2}), thus extracting more discriminative features for object understanding.
To enhances the common pixels while alleviates the ambiguous ones in the feature maps, we uses pixel-level multiplication to get the shared channel-wise features $F^i_{fuse}$: 
\begin{equation}
\label{eq0}
F^i_{fuse} = Concat(\hat{F}^i_{rgb}\otimes \hat{F}^i_{t},\hat{F}^i_{rgb},\hat{F}^i_{t}),
\end{equation}
where $\hat{F}^i_{rgb}$ and $\hat{F}^i_t$ are obtained from $F^i_{rgb}$ and $F^i_t$ through 3$\times$3 convolution operation followed by a batch normalization layer and a ReLU activation function, $Concat(\cdot)$ means the concatenation operation, and $\otimes$ denotes pixel-wise multiplication. 
Then, the shared channel-wise features are fed into the channel attention module~\cite{woo2018cbam}. More specifically, 
\begin{equation}
\label{eq3}
C_{att}^i =\sigma(MLP(P_{max}(F^i_{fuse}))+MLP(P_{avg}(F^i_{fuse})))\otimes F^i_{fuse},
\end{equation}
where $\sigma$ denotes the sigmoid function, and $\otimes$ denotes the multiplication by the dimension broadcast. $P_{max}(\cdot)$ and $P_{avg}(\cdot)$ represent the global max pooling and average pooling operation for each feature map respectively, $MLP(\cdot)$ is a two-layer perceptron.

\textbf{Spatial-Cross Guidance (SCG) module.} 
In this section, we propose a novel SCG module to handle the cross-modal interaction of RGB and thermal images.
Different from most existing interaction strategies~\cite{tu2021multi,huo2021efficient,wang2021cgfnet}, SCG makes the features of two modalities guide each other, and finally supervised by the skeleton label (RGB branch) and contour label (thermal branch) after the correspond decoders, respectively. 
Our design idea comprehensively considers the characteristics of the two modalities (as analyzed in Sec.~\ref{sec:motivation}), so that the spatial consistency can complement each other and improve the robustness of the model under extreme conditions.

Specifically, SCG module firstly deduces the spatial-wise attention features~\cite{woo2018cbam} of the two modalities respectively, and eventually cross-added them to the hierarchical features of the two modalities. The detail operations are defined as follows: 
\begin{equation}
\label{eq4}
S_{att}(F^i) = \sigma (Conv^{7\times7}(Concat(R_{max}(F^i),R_{avg}(F^i)))\otimes F^i,
\end{equation}
where $F^i$ denotes the input feature map, $\sigma$ denotes the sigmoid function. $R_{max}(\cdot)$ and $R_{avg}(\cdot)$ represent the global max pooling and average pooling operation for each point in the feature map along the channel axis respectively, $Concat(\cdot)$ represents a concatenation operation along channel wise, and $Conv^{7\times7}$ represents a convolution operation with filter size $7\times7$.

Finally, the outputted attention interaction maps $Att_{rgb} = \{Att^i_{rgb}|i = 2,3,4,5\}$ and $Att{t} = \{Att^i_{t}|i = 2,3,4,5\}$ are added to the corresponding backbone features of two modalities. So we obtain the low-level interaction features $LF_{rgb} = \{LF^{i}_{rgb}|i = 2,3,4,5\}$ and $LF_{t} = \{LF^{i}_{t}|i = 2,3,4,5\}$ that complement and regulate each other in the channel and spatial dimensions as follows: 
\begin{equation}
\label{eq5}
LF^i_{rgb} = F^i_{rgb} + Att^i_{rgb},\ \  LF^i_{t} = F^i_{t} + Att^i_{t}.
\end{equation}

\begin{figure}
\centerline{\includegraphics[width=\columnwidth]{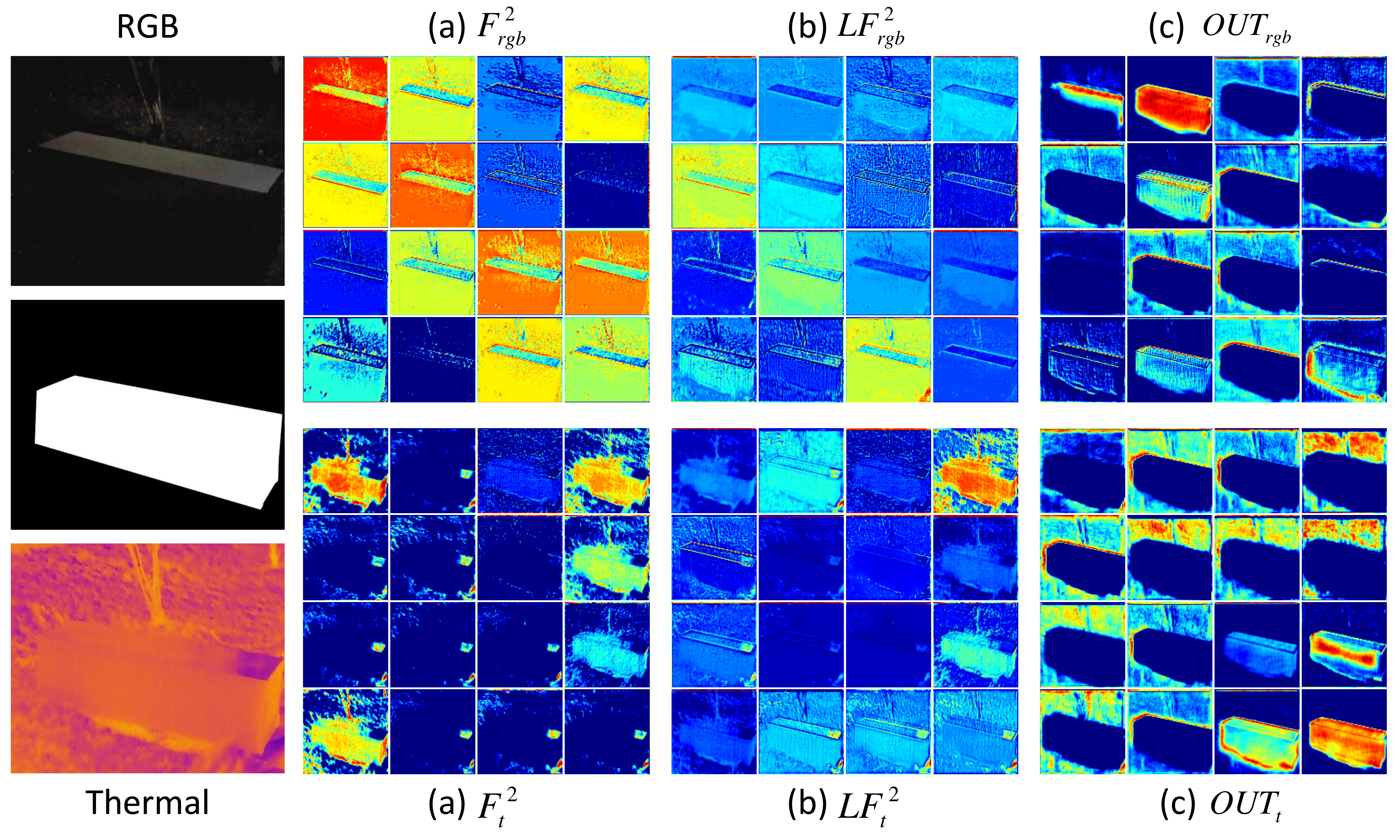}}
\vspace*{-2mm}
\caption{\textbf{Feature visualization in three stages of the proposed mirror complement structure.} The first 16 channels of feature maps are shown in the figure. The first column shows RGB, GT, and Thermal, respectively. (a) $F^2_{rgb}$ and $F^2_{t}$ are the second layers from backbone features. (b) $LF^2_{rgb}$ and $LF^2_{t}$ are the corresponding layer features outputted from attention-based feature interaction module. (c) $OUT_{rgb}$ and $OUT_{t}$ means deep interactive features from SDC module.}
\vspace*{-3mm}
\label{fig6}
\end{figure}

To demonstrate the effect of the proposed attention-based feature interaction module, we show the feature maps outputted by the backbone encoder and attention-based interaction module in challenging scenes with low illumination and thermal cross.
The first column of Fig.~\ref{fig6} provides the input image pair and saliency label. (a) and (b) are the feature maps outputted by the first block of Swin-B and attention-based interaction module respectively. 
(a) $F^2_{rgb}$ and (a) $F^2_{t}$ do not contain a complete salient object due to the low illumination and thermal cross. However, the low-level interaction feature maps (b) $LF^2_{rgb}$ and (b) $LF^2_{t}$ are obviously complementary to each other, and complete object region are mapped on most channels. It confirms that the proposed attention-based feature interaction module can learn the complementary information of the two modalities and model the common saliency region. 
In Sec.~\ref{sec:ablation} we further discuss the impact of variants about the proposed cross-modal attention module on cross-modal saliency modeling. 

\subsection{SDC-based feature fusion module}
In this section, we design a serial multiscale dilated convolution (SDC) based feature fusion module after getting the predictions of the two modalities. 
This module can help the model capture objects of various sizes, especially to improve the segmentation effect of small objects that are difficult to inference. 
Through the interactive iteration of deep semantic features, the proposed model can focus on common salient regions, and obtain more accurate segmentation. 

Specifically, the diverse semantic features of $LF_{rgb}$ and $LF_{t}$ are fused through layer-by-layer aggregated decoders $Dec_{RGB}$ and $Dec_{T}$, thus, the two branches generate corresponding predictions under the supervision of skeleton labels and contour labels, respectively. 
As shown in Fig.~\ref{fig5}, the SDC module takes the concatenated features as inputs to further refine the saliency maps: 
\begin{equation}
\label{eq01}
SDC_{in} = Concat(Dec_{RGB}(LF_{rgb}), Dec_T(LF_{t})),
\end{equation}

Inspired by~\cite{liu2018receptive}, we concatenate a group of dilated convolution with different expansion rates $\gamma = (1,3,5,7)$ to further aggregate the two modalities. Specifically, we set two plain $\gamma \times \gamma$ and 1$\times$1 convolution layers before each dilated convolution layer, and use the maximum pooling after each dilated convolution layer. 
Then these multi-level features outputted by each dilated convolution layer $SDC_{out} = \{SDC^{i}_{out}|i = 2,3,4,5\}$ will be applied with 3$\times$3 convolution layers to obtain two sets of deep interactive features $DF_{rgb} = \{DF^{i}_{rgb}|i = 2,3,4,5\}$ and $DF_{t} = \{DF^{i}_{t}|i = 2,3,4,5\}$, which are suitable for RGB decoder and thermal decoder respectively. The direct superposition method is used to fuse the low-level interaction features $LF_{rgb}$ and $LF_{t}$ with the deep interactive features, and the final saliency map is generated by concatenating these two complementary predictions as follows: 
\begin{equation}
\label{eq02}
Out = Concat(Dec_{RGB}(DF_{rgb}+LF_{rgb}), Dec_T(DF_{t}+LF_{t})).
\end{equation}

\begin{figure*}
\centerline{\includegraphics[width=2.0\columnwidth]{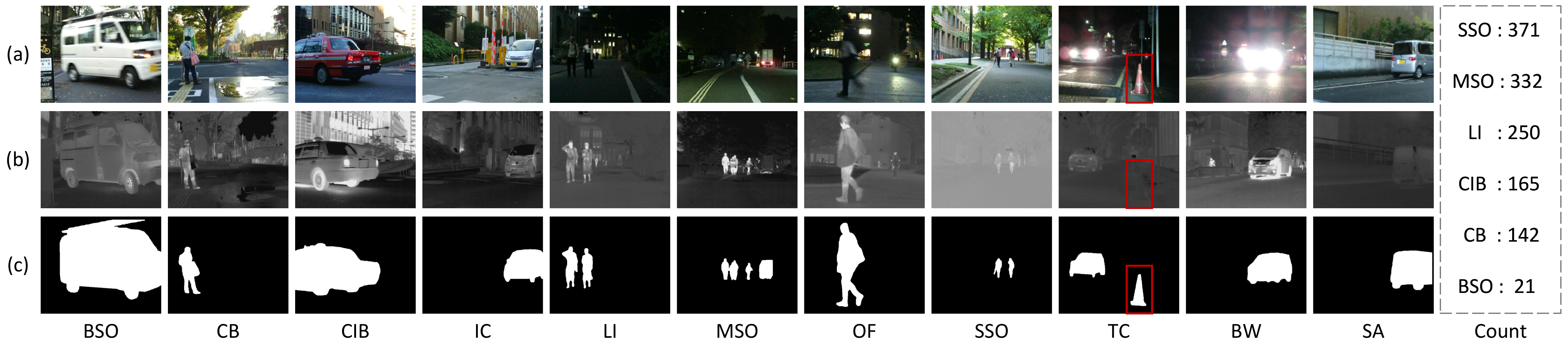}}
\vspace{-3mm}
\caption{\textbf{Sample challenging image pairs and annotated ground truths from our VT723 dataset.} (a) and (b) indicate RGB and thermal image, respectively. (c) is corresponding ground truth of RGB-T image pairs. The last column is the statistics of the number of typical challenging scenes.}
\vspace{-3mm}
\label{fig7}
\end{figure*}

The last column ((c) $OUT_{rgb}$ and $OUT_{t}$) in Fig.~\ref{fig6} shows the feature maps of the two modalities outputted from the SDC module. It proves that the deep interaction makes more channels obtain the complete salient object, while the background clutter is greatly eliminated. 
More ablation experiments can be found in Sec.~\ref{sec:ablation}.

\subsection{RGB-T label decoupling strategy}

Inspired by~\cite{wei2020label}, we design a label decoupling strategy according to the characteristic of RGB-T data. 
Since thermal images usually have distinct edge information (expect for thermal cross scenes) while RGB images provide more details of the salient object, we decompose the original label into a contour map and a skeleton map which are used to supervise thermal and RGB modalities, respectively. Among them, the skeleton map focuses on the main area of the salient object, and the contour map focuses on the outer contour of the salient object. The framework is shown in Fig.~\ref{fig2}.

Specifically, by identifying the Euclidean distance between the pixels belong to salient objects and background, the binary saliency label is converted into a grayscale image (the grayscale value only appears in salient regions).
Then the gray values are normalized to obtain a skeleton map that supervises the RGB modality. Correspondingly, by removing the skeleton map from the original saliency map, the contour map is obtained for supervising the thermal modality.

Finally, the final prediction after fusion module is supervised by the complete label. If both modalities are available, the model could capture the correlation between them and focus its attention on the common salient regions; when suffering extreme lighting conditions or complex thermal environments which results in the failure of any modal, the model could focus on the information-rich branch and accurately segment salient objects with the help of another modal.

\subsection{Loss function}

Based on the proposed RGB-T label decoupling strategy, we obtain three outputs and labels of the RGB branch, the thermal branch, and the fusion module, respectively. So, the total loss $\mathcal{L}$ can be defined as the combination of three losses as follows:
\begin{equation}
\label{eq7}
\mathcal{L} = \ell_{rgb} + \ell_{thermal} + \ell_{fusion},
\end{equation}
where $\ell_{rgb}$, $\ell_{thermal}$ and $\ell_{fusion}$ denote RGB loss, thermal loss and fusion loss, respectively. 

Considering the global structure of the fusion features from RGB and thermal data, we utilize the IoU loss~\cite{mattyus2017deeproadmapper} to measure the similarity and focus on common salient regions.
Moreover, the BCE loss~\cite{de2005tutorial} is utilized to maintain a smooth gradient for all pixels. SSIM loss~\cite{wang2003multiscale} is also introduced into our training. For the RGB and thermal data, SSIM loss helps the optimization to focus on the boundary of salient objects. 
For RGB branch and thermal branch, since skeleton label and contour label are not binary, they cannot be used for IOU loss. So, for $\ell_{rgb}$ and $\ell_{thermal}$, we directly take the sum of BCE loss and SSIM loss. 
Moreover, the hybrid loss $\ell_{fusion}$ is defined as:
\begin{equation}
\label{eq8}
\ell_{fusion} = \ell_{bce} + \ell_{ssim} + \ell_{iou},
\end{equation}
where $\ell_{bce}$, $\ell_{ssim}$ and $\ell_{iou}$ denote BCE loss, IoU loss and SSIM loss, respectively.

\begin{table*}
\caption{\textbf{Performance comparison with state-of-the-art methods on three testing datasets.} The best and second best results are highlighted in red and blue respectively.}
\resizebox{1\linewidth}{!}{ 
    \centering
    \setlength{\tabcolsep}{0.7mm}{
    \begin{tabular}{l|l|cccccc|cccccc|cccccc}
    \toprule[1.5pt]
    \multicolumn{1}{c|}{\multirow{2}{*}{Type}} &
    \multicolumn{1}{c|}{\multirow{2}{*}{Algorithm}} & 
    \multicolumn{6}{c|}{VT5000}                & 
    \multicolumn{6}{c|}{VT1000}                & 
    \multicolumn{6}{c}{VT821}   \\ \cline{3-20} 
    \multicolumn{1}{c|}{}  &
    \multicolumn{1}{c|}{}  & 
    ${\rm F_{\rm avg}\uparrow}$ & ${\rm F_{\rm max}\uparrow}$ & ${\rm F^{\omega}\uparrow}$ & MAE${\downarrow}$ & ${\rm E_{\rm m}\uparrow}$ & ${\rm S_{\rm m}\uparrow}$ &
    ${\rm F_{\rm avg}\uparrow}$ & ${\rm F_{\rm max}\uparrow}$ & ${\rm F^{\omega}\uparrow}$ & MAE${\downarrow}$ & ${\rm E_{\rm m}\uparrow}$ & ${\rm S_{\rm m}\uparrow}$ &
    ${\rm F_{\rm avg}\uparrow}$ & ${\rm F_{\rm max}\uparrow}$ & ${\rm F^{\omega}\uparrow}$ & MAE${\downarrow}$ & ${\rm E_{\rm m}\uparrow}$ & ${\rm S_{\rm m}\uparrow}$ \\
        \midrule[1pt]
        \multicolumn{1}{c|}{\multirow{2}{*}{RGB}} &
        \multicolumn{1}{c|}{GCPANet} & 0.741 & 0.845 & 0.700 & 0.058 & 0.863 & 0.852 & 0.822 & 0.916 & 0.796 & 0.039 & 0.905 & 0.906 & 0.701 & 0.815 & 0.663 & 0.087 & 0.829 & 0.826  \\
        \multicolumn{1}{c|}{}  & \multicolumn{1}{c|}{LDF} & 0.775 & 0.845 & 0.747 & 0.049 & 0.883 & 0.849 & 0.865 & 0.920 & 0.856 & 0.028 & 0.921 & 0.912 & 0.706 & 0.795 & 0.674 & 0.070 & 0.835 & 0.810  \\
        \midrule[1pt]
        \multicolumn{1}{c|}{\multirow{3}{*}{RGB-D}} & \multicolumn{1}{c|}{EFNet} & 0.694 & 0.799 & 0.633 & 0.066 & 0.839 & 0.814 & 0.792 & 0.875 & 0.739 & 0.050 & 0.878 & 0.870 & 0.715 & 0.818 & 0.652 & 0.060 & 0.845 & 0.830  \\
        \multicolumn{1}{c|}{}   & \multicolumn{1}{c|}{RD3D} & 0.796 & 0.866 & 0.728 & 0.048 & 0.903 & 0.864 & 0.844 & 0.919 & 0.812 & 0.035 & 0.917 & 0.905 & 0.719 & 0.809 & 0.653 & 0.078 & 0.841 & 0.819  \\
        \multicolumn{1}{c|}{}   & \multicolumn{1}{c|}{HAINet} & 0.817 & 0.867 & 0.800 & 0.040 & 0.904 & 0.872 & 0.877 & 0.928 & 0.877 & 0.025 & 0.923 & 0.919 & 0.798 & 0.851 & 0.768 & 0.046 & 0.886 & 0.853  \\
        \midrule[1pt]
        \multicolumn{1}{c|}{\multirow{10}{*}{RGB-T}} & 
        \multicolumn{1}{c|}{MTMR} & 0.595 & 0.662 & 0.397 & 0.114 & 0.795 & 0.680 & 0.715 & 0.755 & 0.485 & 0.119 & 0.836 & 0.706 & 0.662 & 0.747 & 0.462 & 0.108 & 0.815 & 0.725  \\
        \multicolumn{1}{c|}{}   & \multicolumn{1}{c|}{M3S-NIR} & 0.575 & 0.644 & 0.327 & 0.168 & 0.780 & 0.652 & 0.717 & 0.769 & 0.463 & 0.145 & 0.827 & 0.726 & 0.734 & 0.780 & 0.407 & 0.140 & 0.859 & 0.723  \\
        \multicolumn{1}{c|}{}   & \multicolumn{1}{c|}{SGDL} & 0.672 & 0.737 & 0.558 & 0.089 & 0.824 & 0.750 & 0.764 & 0.807 & 0.652 & 0.090 & 0.856 & 0.787 & 0.731 & 0.780 & 0.583 & 0.085 & 0.846 & 0.764  \\
        \multicolumn{1}{c|}{}   & \multicolumn{1}{c|}{ADF} & 0.778 & 0.863 & 0.722 & 0.048 & 0.891 & 0.864 & 0.847 & 0.923 & 0.804 & 0.034 & 0.921 & 0.910 & 0.717 & 0.804 & 0.627 & 0.077 & 0.843 & 0.810  \\
        \multicolumn{1}{c|}{}   & \multicolumn{1}{c|}{MIDD} & 0.801 & 0.871 & 0.763 & 0.043 & 0.897 & 0.867 & 0.882 & 0.926 & 0.856 & 0.027 & 0.933 & 0.915 & 0.805 & 0.874 & 0.760 & 0.045 & 0.895 & 0.871  \\
        \multicolumn{1}{c|}{}   &  \multicolumn{1}{c|}{CSRNet} & 
        0.811 & 0.857 & 0.796 & 0.042 & 0.905 & 0.868 & 0.877 & 0.918 & 0.878 & 0.024 & 0.925 & 0.918 & 0.83 & 0.880 & 0.821 & 0.038 & 0.909 & 0.885  \\
        \multicolumn{1}{c|}{}   &  \multicolumn{1}{c|}{ECFFNet} & 0.807 & 0.872 & 0.802 & 0.038 & 0.906 & 0.874 & 
        0.876 & 0.930 & 0.885 & 0.021 & 0.930 & 0.923 & 0.810 & 0.865 & 0.801 & 0.034 & 0.902 & 0.877  \\
        
        \multicolumn{1}{c|}{}   & \multicolumn{1}{c|}{CGFNet} & 0.851 & 0.887 & 0.831 & 0.035 & 0.922 & 0.883 & \textcolor{blue}{\textbf{0.906}} & 0.936 & \textcolor{blue}{\textbf{0.900}} & 0.023 & 0.944 & 0.923 & 0.845 & 0.885 & \textcolor{blue}{\textbf{0.829}} & 0.038 & 0.912 & 0.881  \\
        
        \multicolumn{1}{c|}{}   & \multicolumn{1}{c|}{SwinNet} & \textcolor{blue}{\textbf{0.865}} & \textcolor{blue}{\textbf{0.915}} & 0.846 & \textcolor{blue}{\textbf{0.026}} & \textcolor{blue}{\textbf{0.942}} & \textcolor{blue}{\textbf{0.912}} & 0.896 & \textcolor{blue}{\textbf{0.948}} & 0.894 & 0.018 & 0.947 & \textcolor{blue}{\textbf{0.938}} & \textcolor{blue}{\textbf{0.847}} & \textcolor{blue}{\textbf{0.903}} & 0.818 & 0.030 & 0.926 & \textcolor{blue}{\textbf{0.904}}   \\

        \multicolumn{1}{c|}{}   & \multicolumn{1}{c|}{CAVER} & / & {0.882} & \textcolor{blue}{\textbf{0.849}} & 0.028 & 0.941 & 0.899 & / & {0.939} & \textcolor{blue}{\textbf{0.911}} & \textcolor{blue}{\textbf{0.017}} & \textcolor{blue}{\textbf{0.949}} & \textcolor{blue}{\textbf{0.938}} & / & {0.877} & \textcolor{blue}{\textbf{0.845}} & \textcolor{blue}{\textbf{0.027}} & \textcolor{blue}{\textbf{0.928}} & 0.898  \\
        
        \multicolumn{1}{c|}{}   &  \multicolumn{1}{c|}{Ours} & \textcolor{red}{\textbf{0.892}}  & \textcolor{red}{\textbf{0.926}}  & \textcolor{red}{\textbf{0.891}}  & \textcolor{red}{\textbf{0.021}}  & \textcolor{red}{\textbf{0.953}}  & \textcolor{red}{\textbf{0.924}}  & \textcolor{red}{\textbf{0.920}}  & \textcolor{red}{\textbf{0.957}}  & \textcolor{red}{\textbf{0.929}}  & \textcolor{red}{\textbf{0.013}}  & \textcolor{red}{\textbf{0.955}}  & \textcolor{red}{\textbf{0.948}}  & \textcolor{red}{\textbf{0.878}}  & 
        \textcolor{red}{\textbf{0.925}}  & 
        \textcolor{red}{\textbf{0.881}}  & \textcolor{red}{\textbf{0.021}}  & \textcolor{red}{\textbf{0.938}}  & \textcolor{red}{\textbf{0.923}}  \\
        \bottomrule[1.5pt]
    \end{tabular}}}
    \label{table1}
\end{table*}

\begin{figure*}
\centering
	\subfigtopskip=2pt 
	\subfigbottomskip=2pt 
	\subfigure{
		\includegraphics[width=0.3\linewidth]{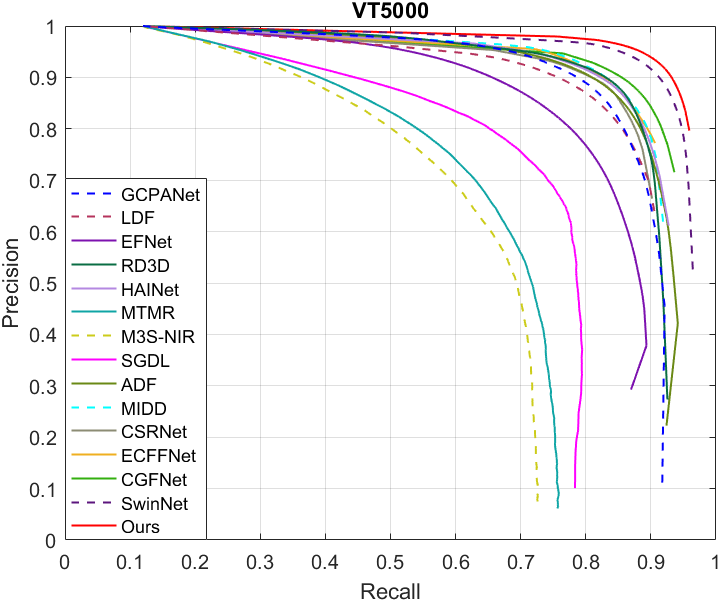}}
	\quad 
	\subfigure{
		\includegraphics[width=0.3\linewidth]{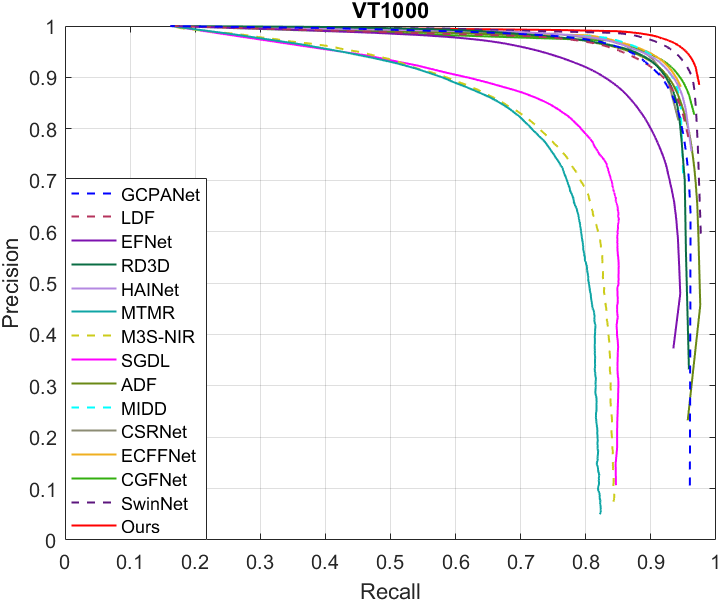}}
	\quad
	\subfigure{
		\includegraphics[width=0.3\linewidth]{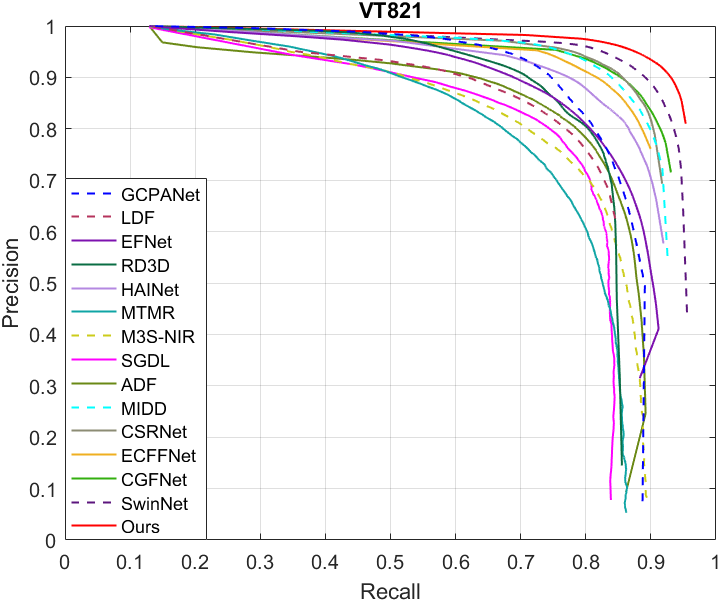}}
	  
	\subfigure{
		\includegraphics[width=0.3\linewidth]{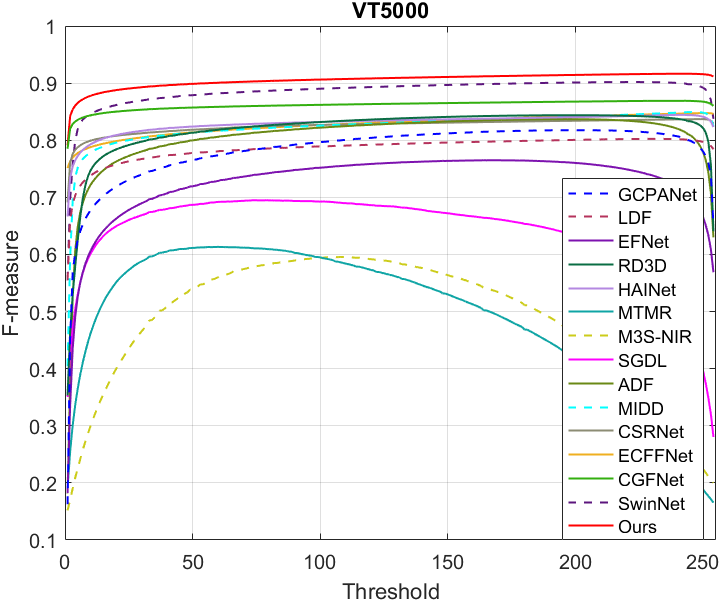}}
	\quad
	\subfigure{
		\includegraphics[width=0.3\linewidth]{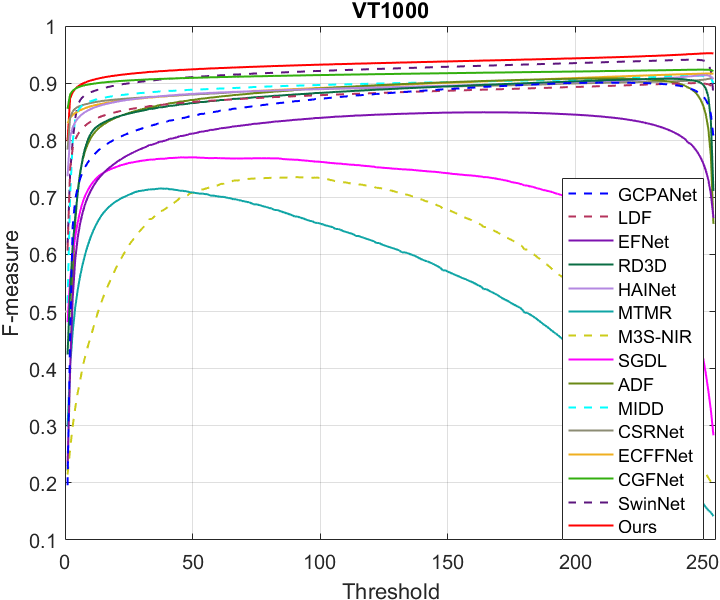}}
	\quad 
	\subfigure{
		\includegraphics[width=0.3\linewidth]{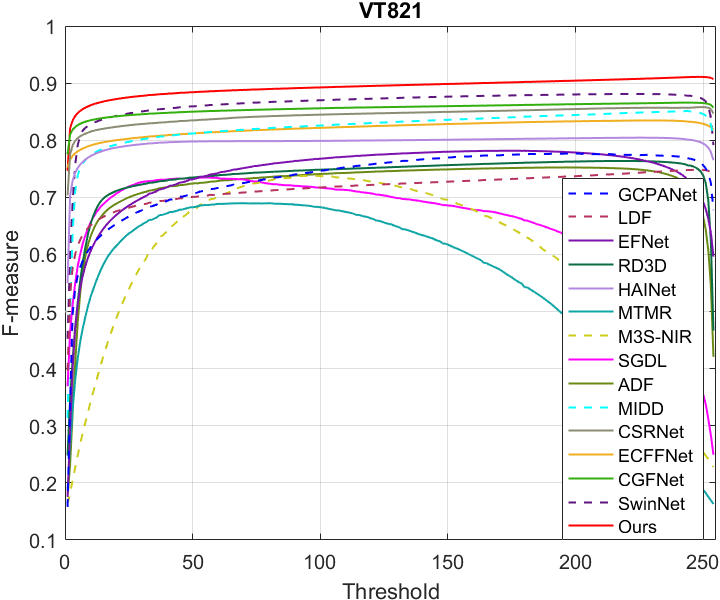}}
	\caption{\textbf{Performance comparison with state-of-the-art SOD methods on VT5000, VT1000, and VT821 datasets.} The first row shows precision-recall curves. The second row shows F-measure curves with different thresholds.}
  \vspace*{-4mm}
  \label{fig8}
\end{figure*}

\section{Experiments}
\subsection{Experimental setup}
\textbf{Public datasets}. 
As mentioned in Sec.~\ref{sec:RGB-T SOD}, there are three RGB-T SOD dataset publicly available, including VT821~\cite{wang2018rgb}, VT1000~\cite{tu2019rgb} and VT5000~\cite{tu2020rgbt}. 
In VT821, the thermal infrared images formed vacant regions during manual registration, and the author also adds noise to some visible images to heighten the challenge of this dataset.
In VT5000, the authors labeled 11 challenging scenes based on factors such as object size, lighting conditions, center deviation, number of prominent objects, and background quality. 
Same as~\cite{tu2021multi,huo2021efficient,zhou2021ecffnet}, 2500 various image pairs in VT5000 are utilized to train our model and the rest image pairs together with VT821 and VT1000 are taken as testing sets.

\begin{table*}
\caption{\textbf{Quantitative comparisons of the proposed method on 11 challenging scenes with ${\rm F_{avg}}$ and MAE metrics.} The best and second best results are highlighted in red and blue respectively.}
\resizebox{1\linewidth}{!}{ 
    \centering
    \setlength{\tabcolsep}{0.7mm}{
    \begin{tabular}{c|c|c|c|c|c|c|c|c|c|c|c}
    \toprule[1.5pt]
    \multicolumn{1}{c|}{\multirow{3}{*}{Algorithm}} & \multicolumn{1}{c}{BSO} & \multicolumn{1}{c}{CB} & \multicolumn{1}{c}{CIB} & \multicolumn{1}{c}{IC} & \multicolumn{1}{c}{LI} & \multicolumn{1}{c}{MSO} & \multicolumn{1}{c}{OF} & \multicolumn{1}{c}{SSO} & \multicolumn{1}{c}{SA} & \multicolumn{1}{c}{TC} & \multicolumn{1}{c}{BW} \\
    \multicolumn{1}{c|}{}  & 
    \multicolumn{1}{c}{887} & \multicolumn{1}{c}{688} & \multicolumn{1}{c}{511} & \multicolumn{1}{c}{591} & \multicolumn{1}{c}{292} & \multicolumn{1}{c}{319} & \multicolumn{1}{c}{179} & \multicolumn{1}{c}{252} & \multicolumn{1}{c}{192} & \multicolumn{1}{c}{450} & 
    \multicolumn{1}{c}{42} \\
    \multicolumn{1}{c|}{}  & 
    \multicolumn{1}{c}{${\rm F_{\rm avg}\uparrow}$/MAE${\downarrow}$} & 
    \multicolumn{1}{c}{${\rm F_{\rm avg}\uparrow}$/MAE${\downarrow}$} & 
    \multicolumn{1}{c}{${\rm F_{\rm avg}\uparrow}$/MAE${\downarrow}$} & 
    \multicolumn{1}{c}{${\rm F_{\rm avg}\uparrow}$/MAE${\downarrow}$} & 
    \multicolumn{1}{c}{${\rm F_{\rm avg}\uparrow}$/MAE${\downarrow}$} & 
    \multicolumn{1}{c}{${\rm F_{\rm avg}\uparrow}$/MAE${\downarrow}$} & 
    \multicolumn{1}{c}{${\rm F_{\rm avg}\uparrow}$/MAE${\downarrow}$} & 
    \multicolumn{1}{c}{${\rm F_{\rm avg}\uparrow}$/MAE${\downarrow}$} & 
    \multicolumn{1}{c}{${\rm F_{\rm avg}\uparrow}$/MAE${\downarrow}$} & 
    \multicolumn{1}{c}{${\rm F_{\rm avg}\uparrow}$/MAE${\downarrow}$} & 
    \multicolumn{1}{c}{${\rm F_{\rm avg}\uparrow}$/MAE${\downarrow}$} \\
    \midrule[1pt]
    
    GCPANet & 0.814/0.073 & 0.745/0.058 & 0.780/0.075 & 0.717/0.069 & 0.726/0.076 & 0.737/0.066 & 0.717/0.072 & 0.480/0.059 & 0.676/0.076 & 0.720/0.062 & 0.677/0.073  \\

    LDF & 0.831/0.070 & 0.776/0.048 & 0.812/0.070 & 0.734/0.061 & 0.795/0.051 & 0.768/0.057 & 0.779/0.049 & 0.615/0.028 & 0.743/0.059 & 0.732/0.057 & 0.715/0.055  \\
    \midrule[1pt]

    EFNet & 0.793/0.093 & 0.656/0.072 & 0.718/0.104 & 0.661/0.079 & 0.726/0.084 & 0.669/0.081 & 0.723/0.069 & 0.390/0.043 & 0.627/0.084 & 0.662/0.070 & 0.638/0.085  \\
    
    RD3D & 0.849/0.066 & 0.797/0.047 & 0.822/0.068 & 0.776/0.055 & 0.819/0.052 & 0.776/0.054 & 0.790/0.053 & 0.623/0.033 & 0.750/0.062 & 0.762/0.060 & 0.751/0.058  \\
    
    HAINet & 0.863/0.059 & 0.815/0.040 & 0.836/0.063 & 0.781/0.050 & 0.823/0.047 & 0.796/0.051 & 0.793/0.049 & 0.681/0.025 & 0.802/0.045 & 0.786/0.043 & 0.761/0.051  \\
    \midrule[1pt]
    
    MTMR & 0.645/0.181 & 0.519/0.121 & 0.541/0.180 & 0.513/0.133 & 0.652/0.131 & 0.571/0.127 & 0.633/0.116 & 0.419/0.035 & 0.567/0.119 & 0.497/0.127 & 0.552/0.115  \\
    
    M3S-NIR & 0.589/0.220 & 0.505/0.185 & 0.512/0.225 & 0.493/0.184 & 0.634/0.181 & 0.551/0.187 & 0.630/0.164 & 0.415/0.135 & 0.531/0.193 & 0.485/0.209 & 0.538/0.168  \\
    
    SGDL & 0.726/0.140 & 0.637/0.092 & 0.647/0.144 & 0.599/0.107 & 0.702/0.112 & 0.640/0.103 & 0.692/0.094 & 0.542/0.027 & 0.577/0.111 & 0.601/0.099 & 0.546/0.112  \\
    
    ADF & 0.843/0.069 & 0.779/0.049 & 0.816/0.073 & 0.755/0.056 & 0.818/0.052 & 0.778/0.054 & 0.780/0.056 & 0.566/0.029 & 0.748/0.058 & 0.739/0.056 & 0.749/0.049  \\
    
    MIDD & 0.861/0.059 & 0.807/0.043 & 0.834/0.063 & 0.766/0.053 & 0.828/0.044 & 0.793/0.051 & 0.804/0.046 & 0.609/0.032 & 0.764/0.054 & 0.766/0.050 & 0.742/0.055  \\
    
    CSRNet & 0.854/0.057 & 0.786/0.046 & 0.808/0.065 & 0.769/0.053 & 0.838/0.043 & 0.788/0.052 & 0.820/0.042 & 0.684/0.028 & 0.766/0.051 & 0.769/0.050 & 0.717/0.055 \\
    
    ECFFNet & 0.867/0.052 & 0.800/0.039 & 0.846/0.054 & 0.776/0.047 & 0.826/0.041 & 0.792/0.043 & 0.798/0.042 & 0.616/0.024 & 0.775/0.047 & 0.778/0.046 & 0.725/0.053  \\

    CGFNet & 0.879/0.057 & 0.850/0.034 & 0.858/0.062 & 0.823/0.042 & 0.866/0.040 & 0.820/0.044 & 0.846/0.041 & \textcolor{blue}{\textbf{0.739}}/0.019 & 0.828/0.044 & 0.827/0.040 & 0.790/0.046  \\

    SwinNet & \textcolor{blue}{\textbf{0.904}}/\textcolor{blue}{\textbf{0.038}} & \textcolor{blue}{\textbf{0.861}}/\textcolor{blue}{\textbf{0.026}} & \textcolor{blue}{\textbf{0.891}}/\textcolor{blue}{\textbf{0.040}} & \textcolor{blue}{\textbf{0.847}}/\textcolor{blue}{\textbf{0.033}} & \textcolor{blue}{\textbf{0.889}}/\textcolor{blue}{\textbf{0.027}} & \textcolor{blue}{\textbf{0.845}}/\textcolor{blue}{\textbf{0.031}} & \textcolor{blue}{\textbf{0.857}}/\textcolor{blue}{\textbf{0.030}} & 0.731/\textcolor{blue}{\textbf{0.011}} & \textcolor{blue}{\textbf{0.849}}/\textcolor{blue}{\textbf{0.030}} & \textcolor{blue}{\textbf{0.849}}/\textcolor{blue}{\textbf{0.028}} & \textcolor{blue}{\textbf{0.803}}/\textcolor{blue}{\textbf{0.038}}   \\
    
    Ours & 
    \textcolor{red}{\textbf{0.919}}/\textcolor{red}{\textbf{0.033}} & \textcolor{red}{\textbf{0.892}}/\textcolor{red}{\textbf{0.020}} & \textcolor{red}{\textbf{0.911}}/\textcolor{red}{\textbf{0.033}} & \textcolor{red}{\textbf{0.876}}/\textcolor{red}{\textbf{0.028}} & 
    \textcolor{red}{\textbf{0.902}}/\textcolor{red}{\textbf{0.024}} & \textcolor{red}{\textbf{0.879}}/\textcolor{red}{\textbf{0.024}} & \textcolor{red}{\textbf{0.879}}/\textcolor{red}{\textbf{0.026}} & \textcolor{red}{\textbf{0.807}}/\textcolor{red}{\textbf{0.008}} & \textcolor{red}{\textbf{0.869}}/\textcolor{red}{\textbf{0.029}} & \textcolor{red}{\textbf{0.883}}/\textcolor{red}{\textbf{0.022}} & \textcolor{red}{\textbf{0.848}}/\textcolor{red}{\textbf{0.023}}  \\
    \bottomrule[1.5pt]
    \end{tabular}}}
    \label{table2}
\end{table*}

\textbf{The proposed challenging dataset VT723}
\label{sec:datasets}
To further validate the robustness of the proposed model under common challenging scenes in real world, we build a more challenging RGB-T SOD dataset VT723 based on a large public semantic segmentation RGB-T dataset~\cite{sun2019rtfnet} used in the autonomous driving domain. We pick image pairs in which the salient objects are significant in at least one of their color and thermal modalities. 
After careful screening, we collectd 723 sets of RGB-thermal image pairs in which 473 are taken during daytime and 250 are taken at night. 
The SOD ground truths are obtained by professional annotators looking at both modalities, selecting common salient regions, and manually marking pixel by pixel on the original segmentation ground truths.

The salient objects in VT723 are mainly vehicles, bicycles, pedestrians, as well as road signs and roadblocks. Since the dataset was captured in an open city street, most of the data have the challenge of multiple salient objects (MSO), complex backgrounds (IC) and center bias (CB), as well as bad weather (BW) challenging scenes. Obviously, data captured at night has low illumination (LI) challenges. In addition, large vehicles typically occupy a large portion of the scene (BSO, size of the big salient object is over 0.26) or even extend beyond the scene boundaries (CIB), while pedestrians and roadblocks usually contain both small salient object (SSO, size of the small salient object is smaller than
0.05) and center bias (CB) challenges. Besides, there are a few scenes in which the object and the background have a similar appearance (SA). 
By looking at the raw image pairs, we also found that the vehicle-to-background (e.g. buildings, roads, trees) contrast in the night thermal images is extremely low (That is, the phenomenon of thermal crossover occurs, TC), which leads to limited information from the thermal images when the RGB images are almost invalid (LI). 

Fig.~\ref{fig7} illustrates the raw images of two modalities and the corresponding ground truth for the above challenging scenes in the proposed VT723 dataset. We summarized the proposed VT723 dataset according to the definitions of typical challenging scenes in~\cite{tu2020rgbt}, the results are listed in the last column of Fig.~\ref{fig7}.

\textbf{Evaluation metrics}. 

In the experiments, we use Precision-Recall (PR) curve, the mean F-measure (${\rm F_{avg}}$)~\cite{achanta2009frequency} , max F-measure (${\rm F_{max}}$), weighted F-measure (${\rm F^{\omega}}$)~\cite{margolin2014evaluate}, mean absolute error (MAE)~\cite{perazzi2012saliency}, E-measure (${\rm E_{m}}$)~\cite{fan2018enhanced}, and S-measure (${\rm S_{m}}$)~\cite{fan2017structure} to evaluate the performance of our method and existing state-of-the-art methods.

\begin{figure*}
\centerline{\includegraphics[width=1.8\columnwidth]{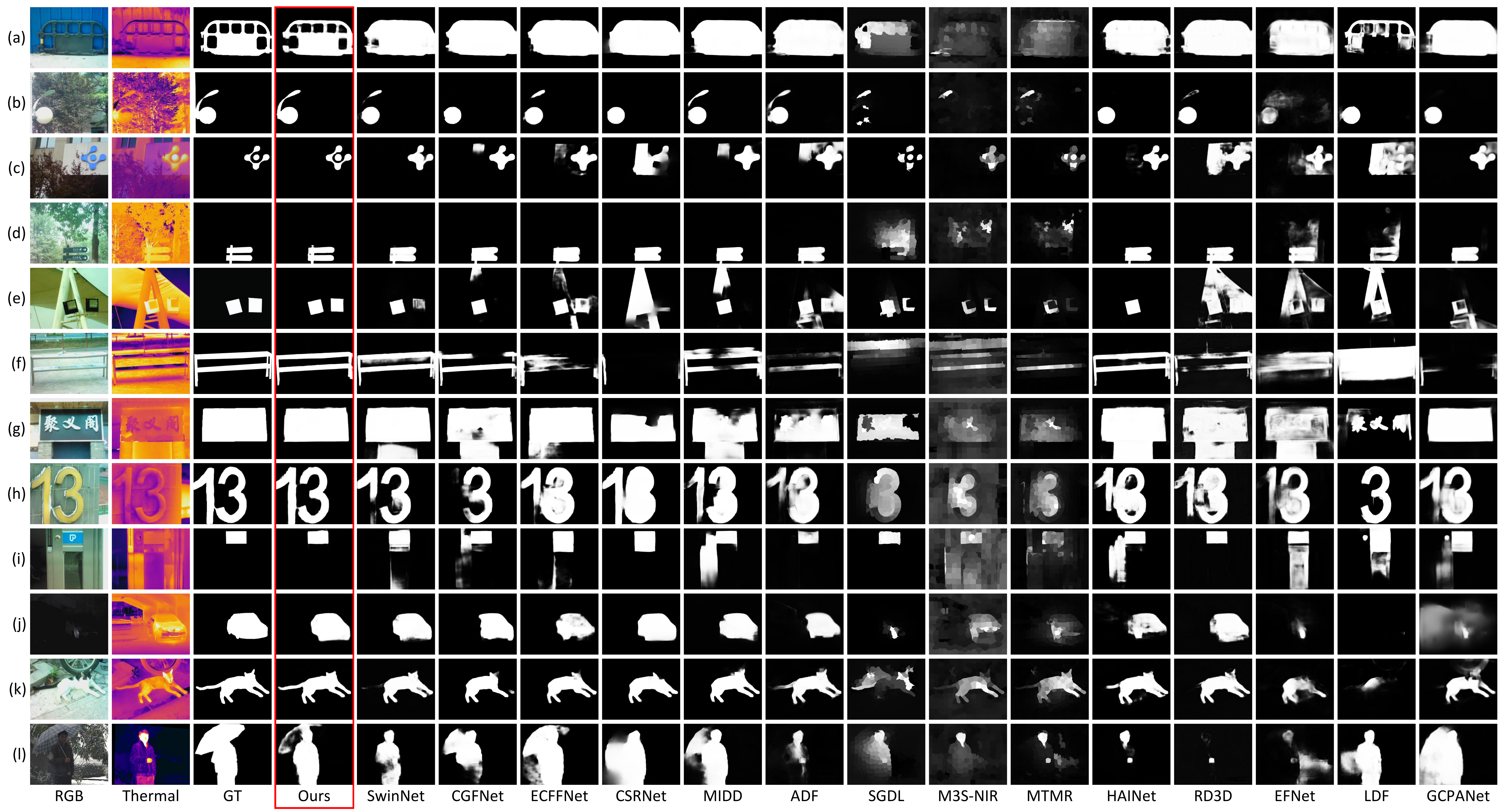}}
\vspace*{-2mm}
\caption{\textbf{Qualitative comparisons with 14 state-of-the-arts methods.} We select 11 RGB-thermal image pairs with diverse challenges to compare the quality of the saliency maps. From the left to right columns are RGB image, thermal image, ground-truth and the results of 15 methods, respectively.}
\label{fig9}
\vspace{-3mm}
\end{figure*}

\textbf{Training details}
We train our model using PyTorch on a single Tesla M40 GPU. The parameters of backbone network are initialized from the Swin-B pretrained on ImageNet22k. The whole network is trained end-to-end by stochastic gradient descent (SGD), momentum and weight decay are set to 0.9 and 0.0005, respectively. The maximum learning rate is 0.005 for backbone of two branches (they do not share weights) and 0.05 for other parts. We train our model 48 epoches and the learning rate increases linearly to maximum in the first half of the iteration cycle and then decreases linearly. In training phase, the batchsize is set to 16, and all the training image pairs of RGB-thermal are augmented using multiple strategies (i.e., random flipping, rotating and border clipping) for data augmentation. We resize the input RGB and thermal images to 384$\times$384 for both the training and test phases. Our model runs at over 8 fps.

\subsection{Comparison with state-of-the-arts on public datasets}
To demonstrate the effectiveness of the proposed method, 15 state-of-the-art SOD methods are introduced to compare as follows: Two deep learning based single-modality SOD methods: GCPANet~\cite{chen2020global} and LDF~\cite{wei2020label}. Three deep learning based RGB-D SOD methods: EFNet~\cite{chen2021ef}, RD3D~\cite{chen2021rgb} and HAINet~\cite{li2021hierarchical}. Three traditional RGB-T SOD methods: MTMR~\cite{wang2018rgb}, M3S-NIR~\cite{tu2019m3s} and SGDL~\cite{tu2019rgb}. And seven deep learning based RGB-T SOD methods: ADF~\cite{tu2020rgbt}, MIDD~\cite{tu2021multi}, CSRNet~\cite{huo2021efficient}, ECFFNet~\cite{zhou2021ecffnet}, CGFNet~\cite{wang2021cgfnet}, SwinNet~\cite{liu2022swinnet}, and CAVER~\cite{pang2021transcmd}. 

All learning-based models are trained on the VT5000 training set (2500 images) described in Sec.~\ref{sec:datasets}.
In experiments, we evaluate the performance based on the saliency maps provided by the original paper (e.g.~\cite{wang2018rgb,tu2019m3s,tu2019rgb,tu2020rgbt,tu2021multi,huo2021efficient,zhou2021ecffnet,liu2022swinnet,pang2021transcmd}). 
For the methods that do not provide saliency maps (e.g.~\cite{chen2020global,wei2020label,chen2021ef,chen2021rgb,li2021hierarchical}), we utilize the code published by the original paper with the recommended parameters to obtain results.
For fair comparisons, RGB-D and single modality SOD methods are retrained in experiments.
Specifically, the early fusion strategy of two modalities~\cite{tu2021multi} is applied to the single modality SOD methods to improve the performance. 
As for the RGB-D SOD methods, the depth map is replaced with the thermal map. 

\begin{table}[t]
\centering
\caption{\textbf{Performance on challenging dataset VT723.} The best and second best results are highlighted in red and blue respectively.}
\resizebox{\columnwidth}{!}{ 
\begin{tabular}{c|cccccc}
\toprule[1.5pt]
Algorithm & ${\rm F_{\rm avg}\uparrow}$ & ${\rm F_{\rm max}\uparrow}$ & ${\rm F^{\omega}\uparrow}$ & MAE${\downarrow}$ & ${\rm E_{\rm m}\uparrow}$ & ${\rm S_{\rm m}\uparrow}$ \\
\midrule[1pt]
GCPANet & 0.501 & 0.582 & 0.434 & 0.073 & 0.746 & 0.704 \\
LDF & 0.453 & 0.565 & 0.391 & 0.108 & 0.699 & 0.654 \\
\midrule[1pt]
EFNet & 0.350 & 0.447 & 0.284 & 0.094 & 0.679 & 0.623 \\
RD3D & 0.567 & 0.632 & 0.392 & 0.073 & 0.813 & 0.702 \\
HAINet & 0.622 & 0.677 & 0.572 & 0.050 & 0.819 & 0.749 \\
\midrule[1pt]
MIDD & 0.583 & 0.701 & 0.502 & 0.066 & 0.772 & 0.756 \\
CGFNet & 0.655 & 0.714 & \textcolor{blue}{\textbf{0.614}} & 0.045 & 0.829 & \textcolor{blue}{\textbf{0.769}} \\
SwinNet & \textcolor{blue}{\textbf{0.693}} & \textcolor{blue}{\textbf{0.734}} & 0.601 & \textcolor{blue}{\textbf{0.037}} & \textcolor{blue}{\textbf{0.878}} & 0.768 \\
Ours & \textcolor{red}{\textbf{0.761}} & \textcolor{red}{\textbf{0.810}} & \textcolor{red}{\textbf{0.724}} & \textcolor{red}{\textbf{0.027}} & \textcolor{red}{\textbf{0.909}} & \textcolor{red}{\textbf{0.833}} \\
\bottomrule[1.5pt]
\end{tabular}}
\label{table3}
\end{table}

\begin{figure}[t]
\centering
	\subfigtopskip=2pt 
	\subfigbottomskip=2pt 
	\subfigure{
		\includegraphics[width=0.48\linewidth]{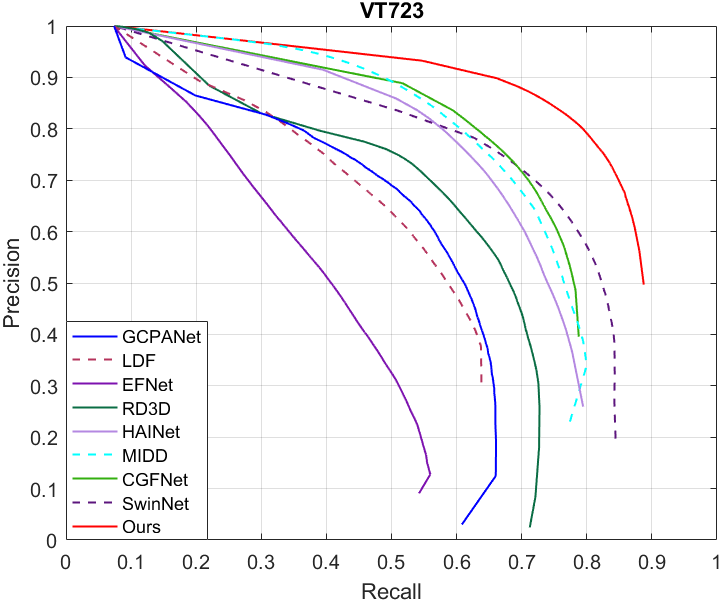}}
	\subfigure{
		\includegraphics[width=0.48\linewidth]{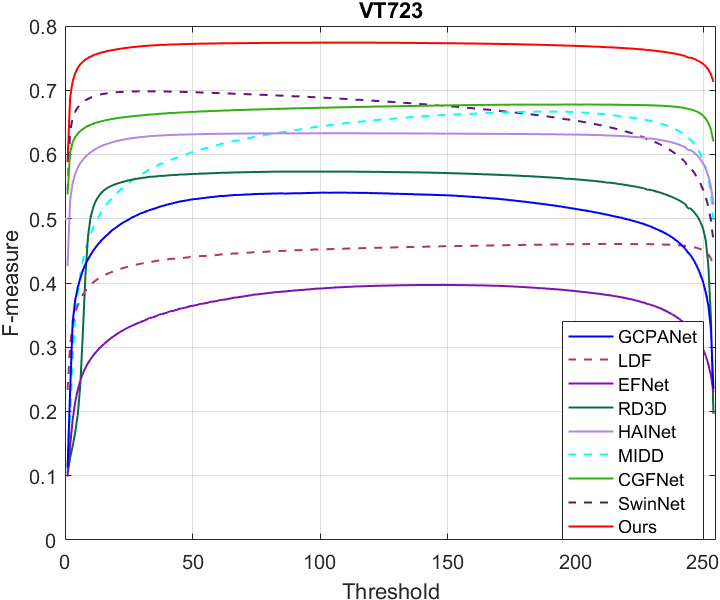}}
	\caption{\textbf{Performance comparison with state-of-the-art methods on VT723 dataset.} The first column shows precision-recall curves. The second column shows F-measure curves with different thresholds.}
   \vspace*{-4mm}
   \label{fig10}
\end{figure}

\textbf{Quantitative evaluation.}
We use the same evaluation code to evaluate the performance of the proposed method and other 15 methods. The quantitative results of CAVER are from the original paper, the author only provides results on partial metrics. As shown in Table~\ref{table1}, the best results are highlighted with red color. 
Obviously, compared with other counterparts on three benchmark datasets, the proposed method outperforms state-of-the-art methods by a large margin. 
In addition, compared with the three traditional methods, the deep learning methods significantly improves the detection performance due to its powerful characterization ability of high-level semantic information. 
The single-modality SOD methods simply takes the integration of the two modalities as input, without explicitly modeling the relationship between them. Therefore, such methods are difficult to obtain satisfactory results in RGB-T SOD task. 
RGB-D SOD methods perform better than above methods since the fusion strategies are more complex. RD3D~\cite{chen2021rgb} is an input fusion model, and the two modalities are gradually fused at the encoder stage. Compared with the single-modality SOD method, it has a significantly improved performance. 
The authors of EFNet~\cite{chen2021ef} and HAINet~\cite{li2021hierarchical} consider their models to be general and prove their validity for RGB-T tasks in the paper. 
EFNet first enhances the depth modality with the prior knowledge learned from the RGB modality and then fuse the RGB and enhanced depth features to obtain final saliency map. 
In the case of poor quality of RGB modality, this method further accumulates error priors, leading to failure to learn effective information of another modality. 
However, in RGB-T task, the low quality RGB image caused by extreme lighting conditions is very common, so the performance of this method is poor. 
HAINet is a typical two-stream structures, and we can find that it also works well on the RGB-T SOD datasets. 
By comparing different methods, we can find that structures which treat the two modalities equally and explicitly model the relationship between them can obtain more accurate saliency maps. 
The precision-recall curves and F-measure curves on three datasets in Fig.~\ref{fig8} also prove the above analysis and conclusion. The curves of the proposed method consistently lie above others.

\begin{figure}[t]
\centerline{\includegraphics[width=1\columnwidth]{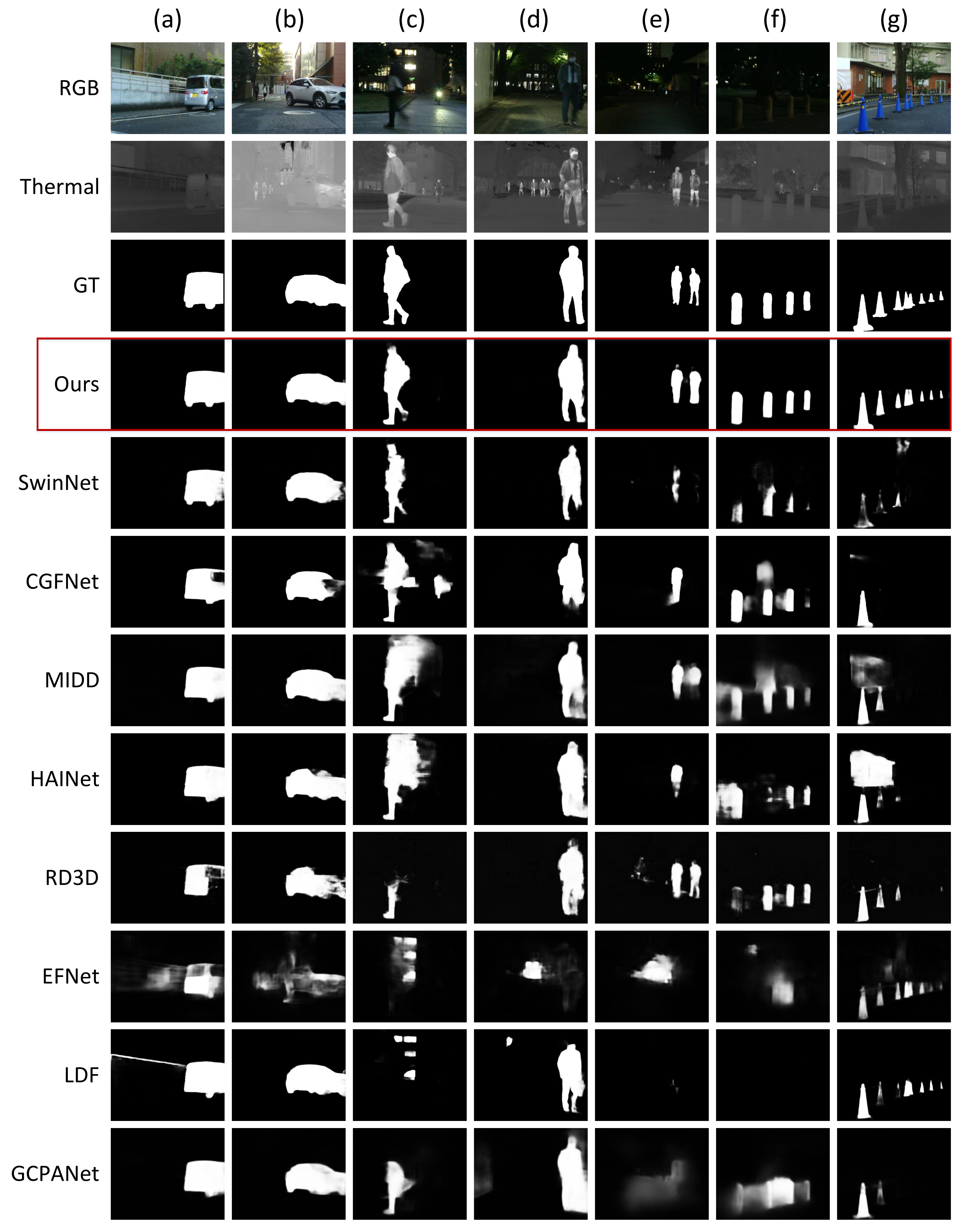}}
\vspace{-1mm}
\caption{\textbf{Qualitative comparisons with state-of-the-arts on VT723 dataset.} We select 9 RGB-thermal image pairs that represent real world challenges to compare the quality of the saliency maps. From top to bottom rows are RGB image, thermal image, ground-truth and the results of 9 methods, respectively.} 
\vspace*{-4mm}
\label{fig11}
\end{figure}

\begin{table*}[t]
\caption{\textbf{Ablation study on different backbones, network architectures and labels on three testing datasets.} The best results are highlighted in red.}
\resizebox{1\linewidth}{!}{ 
    \centering
    \setlength{\tabcolsep}{0.7mm}{
    \begin{tabular}{l|l|cccccc|cccccc|cccccc}
    \toprule[1.5pt]
    \multicolumn{1}{c|}{\multirow{2}{*}{Type}} &
    \multicolumn{1}{c|}{\multirow{2}{*}{Setting}} & 
    \multicolumn{6}{c|}{VT5000}                & 
    \multicolumn{6}{c|}{VT1000}                & 
    \multicolumn{6}{c}{VT821}   \\ \cline{3-20} 
    \multicolumn{1}{c|}{}  &
    \multicolumn{1}{c|}{}  & 
    ${\rm F_{\rm avg}\uparrow}$ & ${\rm F_{\rm max}\uparrow}$ & ${\rm F^{\omega}\uparrow}$ & MAE${\downarrow}$ & ${\rm E_{\rm m}\uparrow}$ & ${\rm S_{\rm m}\uparrow}$ &
    ${\rm F_{\rm avg}\uparrow}$ & ${\rm F_{\rm max}\uparrow}$ & ${\rm F^{\omega}\uparrow}$ & MAE${\downarrow}$ & ${\rm E_{\rm m}\uparrow}$ & ${\rm S_{\rm m}\uparrow}$ &
    ${\rm F_{\rm avg}\uparrow}$ & ${\rm F_{\rm max}\uparrow}$ & ${\rm F^{\omega}\uparrow}$ & MAE${\downarrow}$ & ${\rm E_{\rm m}\uparrow}$ & ${\rm S_{\rm m}\uparrow}$ \\
        \midrule[1pt]
        \multicolumn{1}{c|}{\multirow{6}{*}{Backbone}} &
        \multicolumn{1}{c|}{ResNet-50} & 0.862 & 0.891 & 0.834 & 0.032 & 0.930 & 0.890 & 0.910 & 0.936 & 0.906 & 0.02 & 0.946 & 0.934 & 0.855 & 0.893 & 0.831 & 0.031 & 0.923 & 0.891  \\
        \multicolumn{1}{c|}{}  &
        \multicolumn{1}{c|}{ResNet-101} & 0.845 & 0.891 & 0.835 & 0.032 & 0.923 & 0.895 & 0.901 & 0.944 & 0.907 & 0.018 & 0.941 & 0.937 & 0.835 & 0.889 & 0.832 & 0.036 & 0.910 & 0.893  \\
        \multicolumn{1}{c|}{}  &
        \multicolumn{1}{c|}{Res2Net-50} & 0.844 & 0.891 & 0.833 & 0.032 & 0.922 & 0.894 & 0.899 & 0.940 & 0.903 & 0.019 & 0.939 & 0.935 & 0.836 & 0.892 & 0.832 & 0.031 & 0.915 & 0.893  \\
        \multicolumn{1}{c|}{}  &
        \multicolumn{1}{c|}{PVT-M} & 0.871 & 0.906 & 0.854 & 0.030 & 0.938 & 0.900 & 0.910 & 0.944 & 0.912 & 0.019 & 0.947 & 0.936 & 0.862 & 0.901 & 0.847 & 0.029 & 0.930 & 0.898  \\
        \multicolumn{1}{c|}{}  &
        \multicolumn{1}{c|}{ResNet-50+ViT16} & 0.882 & 0.916 & 0.874 & 0.025 & 0.944 & 0.912 & 0.913 & 0.949 & 0.921 & 0.015 & 0.951 & 0.943 & 0.877 & 0.914 & 0.867 & 0.024 & \textcolor{red}{\textbf{0.938}} & 0.910  \\
        \multicolumn{1}{c|}{}  &
        \multicolumn{1}{c|}{Swin-B, Lower4} & 0.879 & 0.921 & 0.879 & 0.023 & 0.949 & 0.918 & 0.913 & 0.953 & 0.923 & 0.015 & 0.951 & 0.944 & 0.861 & 0.912 & 0.864 & 0.024 & 0.930 & 0.913  \\
        \midrule[1pt]
        \multicolumn{1}{c|}{\multirow{4}{*}{Architecture}} &
        \multicolumn{1}{c|}{Share attention} & 0.891 & 0.924 & 0.888 & 0.022 & 0.953 & 0.922 & 0.915 & 0.951 & 0.925 & 0.014 & \textcolor{red}{\textbf{0.955}} & 0.946 & 0.870 & 0.909 & 0.866 & 0.025 & 0.933 & 0.913  \\
        \multicolumn{1}{c|}{}   &
        \multicolumn{1}{c|}{Cross attention} & 0.888 & 0.923 & 0.886 & 0.022 & 0.952 & 0.923 & 0.916 & 0.954 & 0.925 & 0.014 & 0.953 & 0.946 & 0.872 & 0.919 & 0.874 & 0.022 & 0.935 & 0.919  \\
        \multicolumn{1}{c|}{}   &
        \multicolumn{1}{c|}{Noninteraction attention} & 0.890 & 0.926 & 0.887 & \textcolor{red}{\textbf{0.021}} & \textcolor{red}{\textbf{0.954}} & 0.923 & 0.918 & 0.953 & 0.925 & 0.014 & 0.954 & 0.945 & 0.873 & 0.918 & 0.872 & 0.023 & 0.935 & 0.918  \\
        \multicolumn{1}{c|}{}   &
        \multicolumn{1}{c|}{No SDC} & 0.885 & 0.923 & 0.883 & 0.022 & 0.950 & 0.921 & 0.913 & 0.955 & 0.925 & 0.015 & 0.953 & 0.946 & 0.866 & 0.919 & 0.872 & \textcolor{red}{\textbf{0.021}} & 0.932 & 0.919  \\
        \midrule[1pt]
        \multicolumn{1}{c|}{\multirow{1}{*}{Label}} & 
        \multicolumn{1}{c|}{GT} & 0.884 & 0.925 & 0.884 & 0.022 & 0.950 & 0.922 & 0.912 & 0.952 & 0.924 & 0.015 & 0.951 & 0.945 & 0.862 & 0.919 & 0.868 & 0.023 & 0.932 & 0.918  \\
        \midrule[1pt]
        \multicolumn{1}{c|}{\multirow{1}{*}{-}} & 
        \multicolumn{1}{c|}{Ours} & \textcolor{red}{\textbf{0.892}} & \textcolor{red}{\textbf{0.926}} & \textcolor{red}{\textbf{0.891}} & \textcolor{red}{\textbf{0.021}} & 0.953 & \textcolor{red}{\textbf{0.924}} & \textcolor{red}{\textbf{0.920}} & \textcolor{red}{\textbf{0.957}} & \textcolor{red}{\textbf{0.929}} & \textcolor{red}{\textbf{0.013}} & \textcolor{red}{\textbf{0.955}} & \textcolor{red}{\textbf{0.948}} & \textcolor{red}{\textbf{0.878}} & \textcolor{red}{\textbf{0.925}} & \textcolor{red}{\textbf{0.881}} & \textcolor{red}{\textbf{0.021}} & \textcolor{red}{\textbf{0.938}} & \textcolor{red}{\textbf{0.923}}  \\
        \bottomrule[1.5pt]
    \end{tabular}}}
    \vspace{-2mm}
    \label{table4}
\end{table*}

\textbf{Quantitative evaluation on challenging scenes.} We further conduct a experiment under 11 challenging scenes labeled by VT5000. 
We evaluate the ${\rm F_{avg}}$ and MAE scores of our model as well as 14 state-of-the-art methods. Table~\ref{table2} shows the scores, the numbers of image pairs belonging to the corresponding scenes are also provided. The proposed method performs best on almost all 11 attributes, especially in the following challenging scenes, compared with the second best model SwinNet~\cite{liu2022swinnet}, the ${\rm F_{avg}}$ metric of our method are increase respectively: \textbf{SSO} 9$\%$, \textbf{BW} 5.6$\%$, \textbf{MSO} and \textbf{TC} 4$\%$, which indicating the strong robustness of our model. 
It is worth mentioning that, the outstanding performance in TC (thermal crossover), LI (low illumination) and BW (bad weather) scenes shows the proposed mirror complementary model can flexibly fuse complementary information from both modalities.

\textbf{Qualitative evaluation.} 
Some representative examples of the proposed method and the above 14 SOD methods in a variety of challenging scenes are shown in Fig.~\ref{fig9}. 
Rows (a), (f), and (g) show SA scene in RGB modal, while the objects in rows (a) and (g) contain the challenge of BSO. 
Rows (b), (c), (d), and (i) show challenging scenes of SSO and CB, besides rows (b), (c) and (d) contain IC challenge. 
Rows (c) and (h) show the challenging scenes that contain MSO challenge, and the objects in row (a) also suffer SA challenge. 
The objects shown in rows (a), (c), (d) and (k) contain rich details and complex boundaries, moreover, the different color blocks of the object in RGB modality make it difficult to segment the complete outline. 
The LI scene in row (j) cause the object in the RGB modality invisible, the two single-modality SOD methods cannot capture the salient objects. 
While the scenes in rows (b) and (i), the thermal modality suffers from TC cause the object to almost blend in with the background. 
Since thermal information is not fully utilized, three RGB SOD methods perform not well when the RGB modality contain challenges such as SA and IC (see rows (a) - (f)). 
Compared to SwinNet, our method is robust in most challenging scenes. By effectively combining the complementary features of RGB and thermal, the rich details of the objects can be captured while some ambiguous background regions are suppressed, and the common salient regions of two modalities are highlighted. 
However, when both modalities suffer from severe challenges (e.g., the RGB modality is disturbed by low contrast, and the thermal modality is disturbed by thermal crossover), the performance of our model will also be affected. As shown in row (l) of Fig.~\ref{fig9}, the umbrella behind the person cannot be completely segmented. Nevertheless, thanks to our mirror complementary structure which make the useful features of two modalities effectively extracted and adjusted, the results of our method are still better than other methods.

\begin{figure*}
\centerline{\includegraphics[width=2\columnwidth]{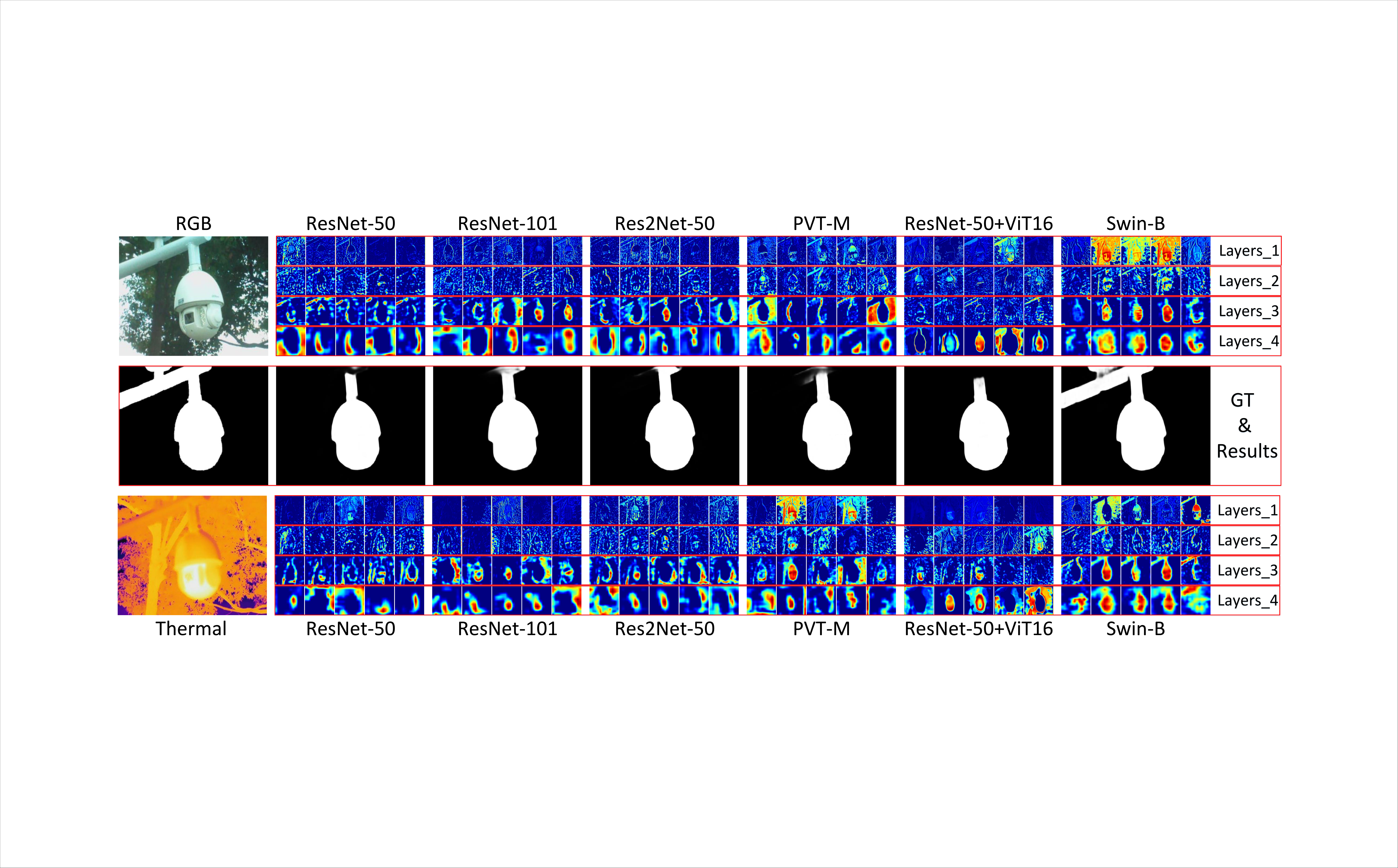}}
\vspace*{-3mm}
\caption{\textbf{Visual comparisons of the hierarchical features of each layer of the backbone networks and the final saliency maps.} 
The first and third rows are RGB and thermal modalities and corresponding hierarchical features respectively, and the middle row are ground truth and the predicted results. 
‘Layers\_*’ represents the outputs of each block of the backbone network (i.e., the hierarchical features used by the proposed model), and we present the feature maps of the first 5 channels for each layer.}
\vspace*{-3mm}
\label{fig12}
\end{figure*}

\subsection{Comparisons on the proposed VT723 dataset}
To further illustrate the superior performance of the proposed method, we conduct experiments on the more challenging dataset VT723. 
Table~\ref{table3} shows the quantitative comparison of the results with eight recent deep-learning based methods which provide the codes: three RGB-T method (SwinNet~\cite{liu2022swinnet}, CGFNet~\cite{wang2021cgfnet} and MIDD~\cite{tu2021multi}), three RGB-D methods (RD3D~\cite{chen2021rgb}, EFNet~\cite{chen2021ef}, and HAINet~\cite{li2021hierarchical}), and two RGB methods (GCPANet~\cite{chen2020global}, and LDF~\cite{wei2020label}). 

Fig.~\ref{fig10} presents the precision-recall curves and F-measure curves on VT723 dataset. It can be seen that compared with other methods, our results have notable improvement on the VT723 dataset. 
We show visualization results of the proposed method and above eight methods in Fig.~\ref{fig11}. 
Our method can highlight the common salient objects of two modalities clearly under various challenging scenes, including SA (column (a)), CB (columns (a), (c) and (e)), SSO (column (e)), MSO (columns (e), (f) and (g)), LI (columns (c)-(f)), TC (columns (a) and (b)) and IC (column (g)). 
In the scenes illustrated in Fig.~\ref{fig11}, center bias location and small size of pedestrian objects, as well as low illumination at nighttime are unavoidable challenges in autonomous driving. 
The results show that the objects of our results are more clear and complete than others. 
The thermal modality plays a crucial role especially in challenging scenes such as severe weather and nighttime, so the model designed for the RGB-T SOD task should make full use of this modality.

\subsection{Ablation studies}
\label{sec:ablation}
In this section, we study the effects of each component of the proposed method on three benchmark datasets. The ablation study contains three parts: backbone ablation, architecture ablation and label ablation.

\textbf{The effectiveness of different backbone}. We compare the effectiveness of mainstream backbones, including ResNet-50~\cite{he2016deep}, ResNet-101~\cite{he2016deep}, Res2Net-50~\cite{gao2019res2net}, Res50+ViT16~\cite{chen2021transunet} and PVT-M~\cite{wang2021pyramid}, and we also report the results when using the low-level four-layer features of Swin-B as in~\cite{liu2022swinnet}. The quantitative comparison results are shown in the first 6 rows of Table~\ref{table4}. The detection performance is significantly improved due to the Swin Transformer taking into account the advantages of both globality and locality. 
In addition, we provide a visual comparison of the hierarchical features of each layer of the backbone networks and the final saliency maps in Fig.~\ref{fig12}. 
The low-level features of the Swin Transformer contain more comprehensive edge and detail information, and its high-level features tend to retain more effective information while filtering out background noise, which is more conducive to downstream dense prediction tasks.

\textbf{The effectiveness of feature interaction and fusion modules}. The middle 4 rows of Table~\ref{table4} illustrates the results of different architecture of the proposed modules. 
‘Share attention’ means that channel attention and spatial attention are successively performed on the shared features $F_{fuse}$, and then two groups of shared attention maps $ATT_{rgb}$ and $ATT_{t}$ with 64 channels are obtained by two sets of 3$\times$3 convolution, which are respectively added to the backbone features $F_{rgb}$ and $F_{t}$ to obtain low-level interaction features $LF_{rgb}$ and $LF_{t}$. 
‘Cross attention’ performs channel attention and spatial attention operations on the backbone features in turn, and the two sets of attention maps obtained are cross-added to the backbone features to obtain low-level interactive features. 
‘Noninteraction attention’ means that the low-level features of the two modalities do not interact, but are the sum of backbone features and their own attention maps: $LF_{rgb}=F_{rgb}+ATT_{rgb}$ and $LF_{t}=F_{t}+ATT_{t}$. And the attention maps are obtained by channel attention and spatial attention operation: $ATT_{rgb}=Sa(Ca(F_{rgb}))$ and $ATT_{t}=Sa(Ca(F_{t}))$. 
The proposed attention-based feature interaction module achieves the best performance on three benchmark datasets, especially on VT821. 
‘No SDC’ means that the serial multiscale dilated convolution
module is not used between the two branches. After getting
the concatenated features from two decoders, we directly use
a set of plain 3$\times$3 convolution layer (instead of the multi-scale
dilated convolution in the SDC module) to obtain the fusion
features of four scales. The proposed SDC-based feature fusion module can significantly improve the performance of the model in each evaluation metrics. 

\textbf{The effectiveness of complementary label pair:} The row of ‘GT’ in Table~\ref{table4} illustrates the results of training without using decoupled labels,
that is both branches and fusion module are supervised by
ground truths. It is clear that the proposed complementary
labels achieve superior qualitative results based on the scores
on three datasets. 

\begin{table}[t]
\caption{\textbf{Comparison of complexity and performance (weighted F-measure metrics on VT 5000 test set). } The model size (num$\_$parameters), FLOPs and $\rm F^{\omega}$ of different models are shown in the Table. The best and second best results are highlighted in red and blue respectively. } 
\resizebox{\columnwidth}{!}{
\centering
\begin{tabular}{l|l|ccc}
\toprule[1.2pt]
Backbone & Algorithm & num$\_$parameters$\downarrow$ & FLOPs$\downarrow$ & ${\rm F^{\omega}\uparrow}$ \\
\midrule[1pt]
\multicolumn{1}{c|}{\multirow{7}{*}{CNN}} &
\multicolumn{1}{c|}{GCPANet} & 67.06M & 26.61G & 0.700 \\
\multicolumn{1}{c|}{} & \multicolumn{1}{c|}{LDF} & \textcolor{red}{\textbf{25.25M}} & \textcolor{red}{\textbf{6.28G}} & 0.747 \\
\cline{2-5} 
\multicolumn{1}{c|}{} & \multicolumn{1}{c|}{EFNet} & \textcolor{blue}{\textbf{38.08M}} & \textcolor{blue}{\textbf{10.10G}} & 0.633 \\
\multicolumn{1}{c|}{} & \multicolumn{1}{c|}{RD3D} & 47.30M & 20.56G & 0.728 \\
\multicolumn{1}{c|}{} & \multicolumn{1}{c|}{HAINet} & 59.80M & 73.55G & 0.800 \\
\cline{2-5} 
\multicolumn{1}{c|}{} & \multicolumn{1}{c|}{MIDD} & 77.59M & 23.25G & 0.763 \\
\multicolumn{1}{c|}{} & \multicolumn{1}{c|}{CGFNet} & 66.38M & 139.97G & \textcolor{blue}{\textbf{0.831}} \\
\multicolumn{1}{c|}{} & \multicolumn{1}{c|}{\textbf{Ours$\_$ResNet50}} & 49.70M & 33.59G & \textcolor{red}{\textbf{0.834}} \\
\hline
\multicolumn{1}{c|}{\multirow{2}{*}{Transformer}} &
\multicolumn{1}{c|}{SwinNet} & \textcolor{blue}{\textbf{199.18M}} & \textcolor{blue}{\textbf{124.14G}} & \textcolor{blue}{\textbf{0.846}} \\
\multicolumn{1}{c|}{} & \multicolumn{1}{c|}{\textbf{Ours$\_$Swin-B}} & \textcolor{red}{\textbf{176.20M}} & \textcolor{red}{\textbf{98.07G}} & \textcolor{red}{\textbf{0.891}} \\
\bottomrule[1.2pt]
\end{tabular}}
\label{table5}
\end{table}

\subsection{Complexity analysis}
Table~\ref{table5} shows the number of parameters and FLOPs (Floating Point Operations) of different algorithms. 
CNN-based single modality SOD methods (e.g., GCPANet~\cite{chen2020global} and LDF~\cite{wei2020label}) usually have a simple network structure because they only process one modal, but their performance is poor in RGB-T SOD task. 
In fact, ‘Ours$\_$ResNet50’ achieves second best in terms of parameter number in CNN-based cross-modal SOD methods (e.g., the RGB-T methods~\cite{tu2021multi,wang2021cgfnet,liu2022swinnet}, the RGB-D methods~\cite{chen2021ef,chen2021rgb,li2021hierarchical}).
Meanwhile, compared with the most recent method CGFNet~\cite{wang2021cgfnet}, ‘Ours$\_$ResNet50’ achieves higher quantitative results with fewer parameters and FLOPs. 
This proves the validity of the proposed mirror complementary structure. 
By combining the mirror structure with Transformer, the performance of ‘Ours$\_$Swin-B’ is improved a large margin than all other methods. 
Compared with the most recent SwinNet~\cite{liu2022swinnet}, the proposed method improves weighted F-measure performance by more than 5$\%$ with fewer parameters and FLOPs, and achiveves SOTA performance. 
In summary, the proposed SOD model can significantly improve the detection performance while reducing the complexity.

\section{Conclusion}

In this paper, we have presented a novel Transformer-based mirror complementary network for RGB-T SOD. 
RGB-T data is significantly different from RGB-D data since their imaging mechanisms are different.
To highlight the common challenge in RGB-T SOD task, we build a novel RGB-T SOD dataset VT723 and make it available to the research community. 
The complementary fusion of cross attention module between low-level Transformer-based features and the SDC interactive between semantics features are proposed to effective capture the details of RGB and thermal modalities. 
To further exploit the characteristics of thermal and RGB images, a set of complementary labels is introduced to separately supervise the RGB images with color and texture information and the thermal images with complete structures and clear edges. 
Expensive experiments demonstrate that the proposed method outperforms state-of-the-art methods under different evaluation metrics, especially improving the performance in challenging scenes. 
Overall, we hope that our work will shed light on the development of more effective RGB-T SOD models.

\ifCLASSOPTIONcaptionsoff
  \newpage
\fi

{\small
\bibliographystyle{IEEEtran}
\bibliography{RGBTbib}
}




%








\end{document}